\documentclass[sigconf, nonacm, oneside]{acmart}
\usepackage{multirow}
\usepackage{colortbl}
\usepackage[ruled,vlined]{algorithm2e}
\usepackage{booktabs}
\usepackage{multirow}
\usepackage{graphicx}
\usepackage{adjustbox}
\usepackage{caption} 
\usepackage{subcaption} 

\AtBeginDocument{%
  }

\setcopyright{acmlicensed}
\copyrightyear{2025}
\acmYear{2025}
\acmDOI{XXXXXXX.XXXXXXX}

\acmISBN{978-1-4503-XXXX-X/18/06}

\begin{document}

\title[Reasoning-based Anomaly Detection Framework]{Reasoning-based Anomaly Detection Framework: A Real-time, Scalable, and Automated Approach to Anomaly Detection Across Domains }

\author{Anupam Panwar}
\authornote{Anupam Panwar is the corresponding author.}
\email{a_panwar@apple.com}
\affiliation{%
  \institution{Apple}
  \city{Cupertino}
  \state{California}
  \country{USA}
}

\author{Himadri Pal}
\email{hpal@apple.com}
\affiliation{%
  \institution{Apple}
  \city{Cupertino}
  \state{California}
  \country{USA}
}

\author{Jiali Chen}
\email{jiali_chen2@apple.com}
\affiliation{%
  \institution{Apple}
  \city{Cupertino}
  \state{California}
  \country{USA}
}

\author{Kyle Cho}
\email{kyle_cho@apple.com}
\affiliation{%
  \institution{Apple}
  \city{Cupertino}
  \state{California}
  \country{USA}
}

\author{Riddick Jiang}
\email{mjiang@apple.com}
\affiliation{%
  \institution{Apple}
  \city{Cupertino}
  \state{California}
  \country{USA}
}

\author{Miao Zhao}
\email{miao_zhao@apple.com}
\affiliation{%
  \institution{Apple}
  \city{Cupertino}
  \state{California}
  \country{USA}
}

\author{Rajiv Krishnamurthy}
\email{rajiv_krishnamurthy@apple.com}
\affiliation{%
  \institution{Apple}
  \city{Cupertino}
  \state{California}
  \country{USA}
}

\renewcommand{\shortauthors}{Panwar et al.}

\begin{abstract}

Detecting anomalies in large, distributed systems presents several challenges. The first challenge arises from the sheer volume of data that needs to be processed. Flagging anomalies in a high-throughput environment calls for a careful consideration of both algorithm and system design. The second challenge comes from the heterogeneity of time-series datasets that leverage such a system in production. In practice, anomaly detection systems are rarely deployed for a single use case. Typically, there are several metrics to monitor, often across several domains (e.g. engineering, business and operations). A one-size-fits-all approach rarely works, so these systems need to be fine-tuned for every application - this is often done manually. The third challenge comes from the fact that determining the root-cause of anomalies in such settings is akin to finding a needle in a haystack. Identifying (in real time) a time-series dataset that is associated causally with the anomalous time-series data is a very difficult problem. In this paper, we describe a unified framework	 that addresses these challenges. Reasoning based Anomaly Detection Framework (RADF) is designed to perform real time anomaly detection on very large datasets. This framework employs a novel technique (mSelect) that automates the process of algorithm selection and hyper-parameter tuning for each use case. Finally, it incorporates a post-detection capability that allows for faster triaging and root-cause determination. Our extensive experiments demonstrate that RADF, powered by mSelect, surpasses state-of-the-art anomaly detection models in AUC performance for 5 out of 9 public benchmarking datasets. RADF achieved an AUC of over 0.85 for 7 out of 9 datasets, a distinction unmatched by any other state-of-the-art model. 

\end{abstract}



\maketitle


\section{Introduction}

Anomaly Detection is the process of identifying data points that deviate significantly from expected values in a dataset. These deviations (henceforth referred to as anomalies) are often of special interest and require timely intervention. For instance, a large transaction (say \$1000) in a credit card dataset where most transactions are small (say less than \$10) could indicate fraud and may trigger actions such as disabling the card. Traditional anomaly detection methods, which are predominantly rule-based, rely on manually defined thresholds and heuristics to identify anomalies. While these approaches may suffice initially, they fall short in the long run. For example, they struggle with noisy datasets, require frequent manual updates to thresholds due to changing trends, lack flexibility to adapt to complex or multi-dimensional data distributions, and are prone to missing subtle patterns or relationships in the data.

This motivates the need to apply specialized algorithms to detect anomalies in large scale, distributed systems. While there are several statistical and ML based algorithms that can detect anomalies, the candidate set quickly diminishes when these need to be applied at very large scale, in real time. This is because these algorithms need to be computationally efficient and should have low latency in response time. Statistical methods, such as Robust Seasonal Extreme Studentized Deviate (ESD) \cite{hochenbaum2017automatic}, detect anomalies by accounting for seasonality and extreme deviations. Unsupervised models such as LSTM-VAE \cite{Park_2018} use LSTMnetworks and Variational Autoencoders to learn normal patterns without labeled data. Semi-supervised methods, such as ACVAE \cite{9844802}, leverage adversarial training to enhance detection using primarily normal data. Supervised techniques such as Opprentice \cite{10.1145/2815675.2815679} rely on labeled datasets for precise anomaly detection, especially when sufficient labeled anomalies are available. These approaches significantly improve anomaly detection performance and scalability.
However, they still have limitations, such as the need for labeled data in some methods, and others requiring manual model and parameter selection for each time series.

Real-world time series data involve multiple metrics and dimensions, significantly complicating anomaly detection. Identifying an optimal model for every combination of these factors is challenging, as time series characteristics—such as stability and seasonality—vary widely across domains. For example, cloud computing operational data often behaves differently from payment transaction data. In large-scale online systems, which may involve hundreds of thousands of dimensional combinations, manually selecting models and tuning parameters is impractical due to high overhead. This issue is especially pronounced in unsupervised models, where the lack of ground truth makes it difficult to determine or validate the best model.
To address these challenges, we propose two key criteria for automating unsupervised anomaly detection in large-scale systems: \textbf{Automatic Model Selection (AMS)}: This ensures model scalability by automatically choosing the best-fitting algorithms for hundreds of thousands of metric and dimension combinations, eliminating the need for manual intervention. AMS enables efficient monitoring of large numbers of Key Performance Indicators (KPIs). \textbf{Domain Knowledge Abstraction}: By abstracting the domain-specific knowledge required for anomaly detection, end users, such as engineering teams, can easily onboard use cases without needing extensive expertise. This lowers the barrier to entry and accelerates adoption.

In real-world applications, root cause analysis (RCA) is traditionally treated as a module that is separate from anomaly detection, where the goal is to identify potential root causes given the detected anomalous metrics by analyzing the dependencies between the monitored metrics. Because RCA requires knowing which metric is anomalous, univariate (instead of multivariate) time series anomaly detection algorithms are mostly applied to detect anomalies, and then RCA analyzes system/service graphs obtained via domain knowledge or observed data to determine root causes. Both univariate and multivariate algorithms have drawbacks and cannot be integrated with RCA seamlessly.

Another major challenge is providing reasoning once anomalies are detected. In practical applications, anomalies have limited value without the necessary context or explanation. Most existing methods offer interpretability for specific models, such as DA-VAE \cite{10.1145/3589335.3651492}, but lack generalization across all anomaly detection techniques. This limitation is particularly problematic in real-world complex KPI time series data, where two primary issues arise:

\begin{itemize}
 \item \textbf{Model-Agnostic Root Cause Analysis (RCA)}: While some recent research offers explanations tied to specific models, they fail to generalize across all anomaly detection methods. Furthermore, RCA should operate across both dimensions and metrics. For example, consider a KPI like transaction\_amount for a company operating in multiple countries (dimensions). If an anomaly (potential fraud) is detected on a specific date, determining which country triggered the anomaly exemplifies cross-dimension RCA. Similarly, if we seek to identify the cause across different metrics (e.g., correlating transaction\_amount with customer\_count), it involves cross-metric RCA. Addressing these RCA challenges is crucial for actionable insights in anomaly detection.
 \item \textbf{Real-time Reasoning for Anomaly Detection}: In real-world systems, detecting anomalies alone is insufficient and providing the context or reasoning behind these anomalies is critical for actionable insights. This is particularly true for systems operating in diverse environments where rapid and accurate root cause analysis (RCA) is necessary to prevent downtime or mitigate risks \cite{wu2020developing}. Key frameworks, such as DA-VAE \cite{10.1145/3589335.3651492}, focus on model-specific explanations, but broader generalization is needed for different anomaly detection techniques. Another challenge is the development of unsupervised models capable of offering real-time explanations without requiring labeled data, a crucial need in scenarios where obtaining ground truth is impractical.
\end{itemize}

In this paper, we introduce RADF, a Reasoning-based Anomaly Detection Framework designed to offer scalable anomaly detection with real-time root cause analysis (RCA). To the best of our knowledge, RADF is the first framework that integrates end-to-end model selection for unsupervised anomaly detection and provides reasoning through causal and correlation analysis.
For model selection, we evaluated RADF on 11 datasets containing 205 time series and 40,000 data points. Of these, 124 time series exhibited stable patterns, 72 displayed instability, and 9 demonstrated a trend. A stable time series exhibits consistent patterns or behaviors over time with minimal fluctuations, such as daily sales of a product with steady demand. An unstable time series is characterized by irregular or erratic fluctuations, often driven by high variability or noise, like social media activity spikes during viral events. In contrast, a trend-based time series shows a clear upward or downward trajectory over time, reflecting long-term changes, such as a steady increase in subscribers to a streaming service over months. All data was labeled by human evaluators. To evaluate the effectiveness of our novel model selection algorithm in an unsupervised setting, we employed standard metrics such as Precision, Recall, and F1-score. Additionally, we developed an RCA module based on causal and correlation analysis, which enables the identification of root causes with quantifiable contributions from related time series. Notably, labels were used only for evaluation and not for training.
RADF also offers a configuration-driven framework to deploy anomaly detection pipelines at scale. It supports 19 anomaly detection algorithms, two change-point detection methods, three decomposers, and three smoothing algorithms.

Our contributions are summarized as follows:
\begin{itemize}
\item RADF introduces a scalable anomaly detection framework that automates model selection for unsupervised anomaly detection in unlabeled time series data. By benchmarking performance across 11 datasets with varying patterns, RADF ensures adaptability and optimal results, evaluated using Precision, Recall, and F1-score metrics.

\item RADF includes a causality and correlation based Root Cause Analysis (RCA) module, which enhances anomaly interpretability by identifying potential root causes through cross-metric and cross-dimension analysis. This provides actionable insights, helping users understand the context behind detected anomalies.

\item RADF’s config-driven design streamlines large-scale deployment, processing terabyte-scale data efficiently. It supports 19 anomaly detection models alongside various auxiliary algorithms for change-point detection, decomposition, and smoothing, minimizing manual intervention and simplifying operational workflows.

\item RADF offers a robust, enterprise-grade comprehensive framework that extends beyond traditional anomaly detection by incorporating automated model selection, distributed computing, and intelligent reasoning, making it well-suited for real-world enterprise applications.
\end{itemize}

These contributions position RADF as a comprehensive solution for scalable, interpretable anomaly detection in diverse real-world scenarios.


\section{Related Work}
In this section, we first examine the existing methods for automatic model selection or hyperparameter tuning algorithms, then we review time series Root Cause Analysis (RCA) algorithms. Afterwards, we will give an overview of distributed systems, and lastly, we analyze the existing anomaly detection frameworks. This comprehensive review provides the foundation for understanding RADF's novel contributions and contextualizes our comparative evaluation presented in Section~\ref{experiments}.

\subsection{Automatic Model Selection and Hyperparameter Tuning}
In the context of anomaly detection, particularly for unsupervised learning, model selection and hyperparameter tuning are challenging due to the absence of labeled data. Traditional methods like grid search and random search remain popular but are computationally inefficient for high-dimensional datasets commonly encountered in time series anomaly detection~\cite{bergstra2012random}. These approaches often struggle to identify optimal configurations in dynamic and diverse data scenarios. Bayesian optimization has emerged as a promising technique for unsupervised anomaly detection, building surrogate models of the objective function to iteratively refine hyperparameters~\cite{snoek2012practical}. However, the method's reliance on well-defined evaluation metrics can limit its applicability in purely unsupervised settings where ground truth is unavailable. In addition to hyperparameter tuning, meta-learning has been explored to automate algorithm selection by transferring knowledge from supervised neighbor tasks to unsupervised novel tasks~\cite{zhao2021automatic, singh2022meta, zhao2022toward, navarro2023meta}. This approach enables a more robust search for optimal configurations by leveraging prior learning experiences.

Beyond that, multiple researchers have explored building anomaly detection ensembles. Methods that aggregate outputs from all detectors, such as those proposed by Lazarevic and Kumar~\cite{lazarevic2005feature}, do not account for the accuracy of individual detectors, which can degrade overall performance when inaccurate detectors are included. Alternatively, diversity-based methods, as explored by Schubert et al.~\cite{schubert2012evaluation}, Zimek et al.~\cite{zimek2014ensembles} and Klementiev et al.~\cite{klementiev2009unsupervised}, aim to increase error independence among detectors but risk incorporating unreliable results for the sake of diversity. Moreover, Rayana et al.\cite{rayana2016less} and Ying et al.~\cite{ying2020automated} proposed automated approaches for ensemble anomaly detection algorithms by understanding the timeseries with the feature extraction and select algorithms based on that.

\subsection{Root Cause Analysis}
Root Cause Analysis (RCA) methodologies have seen significant advancements, particularly in the context of time-series data. Early approaches like PWGC~\cite{granger1969investigating} and MVGC~\cite{geweke1982measurement} provided statistical tools for identifying causal relationships but were limited by assumptions of linearity and stationarity. Recent extensions, such as the TCDF~\cite{nauta2019causal}, address non-linear dependencies using deep learning techniques. Constraint-based methods, including PCMCI~\cite{runge2020discovering}, oCSE (Optimal Causation Entropy)~\cite{sun2015causal}, and tsFCI~\cite{malinsky2018causal}, exploit conditional independencies to infer causal structures, while more advanced methods like ANLTSM and SVAR-FCI~\cite{eichler2013causal} enhance scalability and accuracy. Noise-based approaches, such as VarLiNGAM~\cite{hyvarinen2010estimation} and TiMINo~\cite{peters2016causal}, expand the scope by leveraging statistical independence assumptions. Meanwhile, score-based techniques like DYNOTEARS~\cite{pamfil2020dynotears} integrate optimization frameworks to infer causal graphs effectively. Moreover, due to the topology nature of causality, graphical neural networks(GNN)~\cite{jiang2023graph} have also been explored to analyze causality. Despite these advancements, challenges persist in handling context-specific requirements, latent confounders, and computational complexity, highlighting the need for adaptable, scalable and robust system to support RCA algorithms on real world, large scale applications.


\subsection{Anomaly Detection Frameworks}
End-to-end anomaly detection solutions are pivotal for identifying and mitigating unexpected behaviors in complex systems. These solutions integrate data collection, preprocessing, model training, and real-time inference into a seamless workflow. For example, ADBench~\cite{han2022adbench}, ANORMALib~\cite{akcay2022anomalib} and AdaTime~\cite{adatime} offers comprehensive collection of Anomaly Detection algorithms and the most popular datasets for benchmarking and researching; ADecimo~\cite{boniolMSAD2023} provides a pipeline that not only offers the algorithms but also the model selection functionality; EGADS~\cite{laptev2015generic} and Tods~\cite{Lai_Zha_Wang_Xu_Zhao_Kumar_Chen_Zumkhawaka_Wan_Martinez_Hu_2021} offer scalable end-to-end anomaly detection system for scalability. To the best of our knowledge, our work is the first to consolidate Automated Ensemble Model and Hyperparameter Selection, Root Cause Analysis and Anomaly Detection together for a large scale dataset.


\section{RADF: Reasoning-based Anomaly Detection Framework}

In this section, we introduce RADF from the aspects of architecture and workflow. Then we provide details on all the modules that are a part of the framework.

\subsection{Overview}
\label{subsec:radf_overview}

RADF is a comprehensive novel framework designed to autonomously construct, execute, and monitor anomaly detection pipelines for both batch and real-time data processing. This framework facilitates the seamless integration and deployment of various anomaly detection algorithms through a simple configuration interface, enabling the efficient execution of time series analysis \cite{10.1145/1541880.1541882}. RADF supports both univariate and multivariate time series datasets, making it a versatile tool for different anomaly detection tasks.
RADF consists of two main components: the Core Library and the Orchestrator. The Core Library includes a rich set of anomaly detection algorithms, model selection techniques, change point detection methods, smoothing and decomposing functions, and root cause analysis tools. These components are designed to streamline the process of building robust anomaly detection pipelines, leveraging cutting-edge research in time series analysis \cite{10.1016/j.jnca.2015.11.016, dau2019ucrtimeseriesarchive}.

\subsubsection{\textbf{Core Library}}
    The Core Library serves as the backbone of RADF, encompassing a wide range of specialized algorithms designed for diverse anomaly detection tasks. Currently, the Core Library supports a total of 33 algorithms, including those for anomaly detection, change point detection, smoothing, and more, as detailed in Table~\ref{tab:algorithm_table}. These include:

\begin{table}[t]
\captionsetup{font=small} 
\centering
\caption{Overview of Algorithms and Characteristics}
\label{tab:algorithm_table}
\footnotesize 
\setlength{\tabcolsep}{2pt} 
\renewcommand{\arraystretch}{0.65} 
\resizebox{0.85\columnwidth}{!}{
\begin{tabular}{cccc}
\toprule
\textbf{Algorithm Type} & \textbf{Algorithm Category} & \textbf{Support Multivariate} & \textbf{Total Algorithms} \\ 
\midrule
\multirow{7}{*}{Anomaly Detector} 
 & \multirow{2}{*}{Statistics Based} & Yes & 1 \\ 
 &                                      & No & 10 \\ 
\cmidrule{2-4}
 & \multirow{2}{*}{Machine Learning Based} & Yes & 1 \\ 
 &                                          & No & 2 \\ 
\cmidrule{2-4}
 & Signal Processing Based & No & 1 \\ 
\cmidrule{2-4}
 & Rule Based Approach & No & 2 \\ 
\cmidrule{2-4}
 & Ensemble Learning & Yes & 1 \\ 
\midrule
\multirow{2}{*}{Change Point Detector} 
 & \multirow{2}{*}{Statistics Based} & Yes & 1 \\ 
 &                                      & No & 1 \\ 
\midrule
\multirow{3}{*}{Smoother} 
 & Rolling Window & Yes & 2 \\ 
\cmidrule{2-4}
 & Cyclic Subseries & Yes & 1 \\ 
\cmidrule{2-4}
 & Ensemble Learning Method & Yes & 1 \\ 
\midrule
\multirow{3}{*}{Decomposer} 
 & Simple Decomposer & Yes & 1 \\ 
\cmidrule{2-4}
 & RPCA Decomposer & Yes & 1 \\ 
\cmidrule{2-4}
 & STL Decomposer & Yes & 1 \\ 
\midrule
\multirow{3}{*}{Root Cause Analyzer} 
 & Correlation Coefficient & No & 1 \\ 
\cmidrule{2-4}
 & Distance Based & No & 2 \\ 
\cmidrule{2-4}
 & Prediction Based & No & 1 \\ 
\midrule
\multirow{2}{*}{Model Selector} 
 & Without Label & No & 1 \\ 
\cmidrule{2-4}
 & With Label & No & 1 \\ 
\bottomrule
\end{tabular}%
}

\end{table}
    
 \begin{itemize}
 
        \item Anomaly Detection Algorithms: The framework supports multiple statistical and machine learning techniques for anomaly detection, drawing on established methods and frameworks \cite{10.1145/1541880.1541882}. The Core library currently supports 19 anomaly detection algorithms, catering to both univariate and multivariate time series data. RADF supports algorithms such as Long Short-Term Memory Variational Autoencoder (LSTM VAE) \cite{Zhang2019TimeSA}, Enhanced Isolation Forest \cite{10.1016/j.patcog.2021.108115}, and others. A general equation might look like this: 

\begin{align}
A(x) &= 
\begin{cases} 
1, & \text{if } S(x) > \tau, \\
0, & \text{otherwise,}.
\end{cases} & \label{eq:anomaly_indicator} \\
\text{where:} & \nonumber \\
\quad & A(x): \text{ Anomaly indicator function.} & \nonumber \\
\quad & S(x): \text{ Anomaly score function.} & \nonumber \\
\quad & \tau: \text{ Threshold value.} & \nonumber
\end{align}

        \item Change Point Detection Algorithms: These algorithms are designed to pinpoint shifts in the underlying distribution or trend of a time series, known as change points. Identifying such points is critical for understanding structural breaks or significant transitions in data, with applications in system health monitoring and fraud detection \cite{Killick_2012}.
        \item Smoothing Algorithms: Smoothing techniques mitigate the effects of noise and irregular fluctuations in time series data, enhancing the visibility of trends and patterns. This improves the accuracy of subsequent anomaly detection processes. Common smoothing methods include moving averages, Gaussian smoothing, and exponential smoothing.
        \item Decomposition Algorithms: Time series decomposition separates a series into components such as trend, seasonality, and residuals. This separation facilitates individual component analysis, improving anomaly detection precision in residuals. The technique enhances interpretability and the performance of anomaly detection models \cite{Zhang2019TimeSA}.
        \item Root Cause Analysis (RCA) Algorithms: RCA algorithms quantifies relationships between multiple time series, assessing how variations in one influence the target series. This analysis is crucial in multivariate contexts, where interactions among variables play a significant role.
        \item Model Selection Algorithms: Selecting the most appropriate model and parameters is a critical step in machine learning and time series analysis. RADF employs techniques such as cross-validation, hyperparameter optimization, and performance evaluation to recommend the best model configuration. Ground truth annotations, when available, further enhance the selection process.
        
\end{itemize}

\subsubsection{\textbf{Orchestrator}}
\begin{figure}[htbp]
    \centering
    \includegraphics[width=0.9\linewidth]{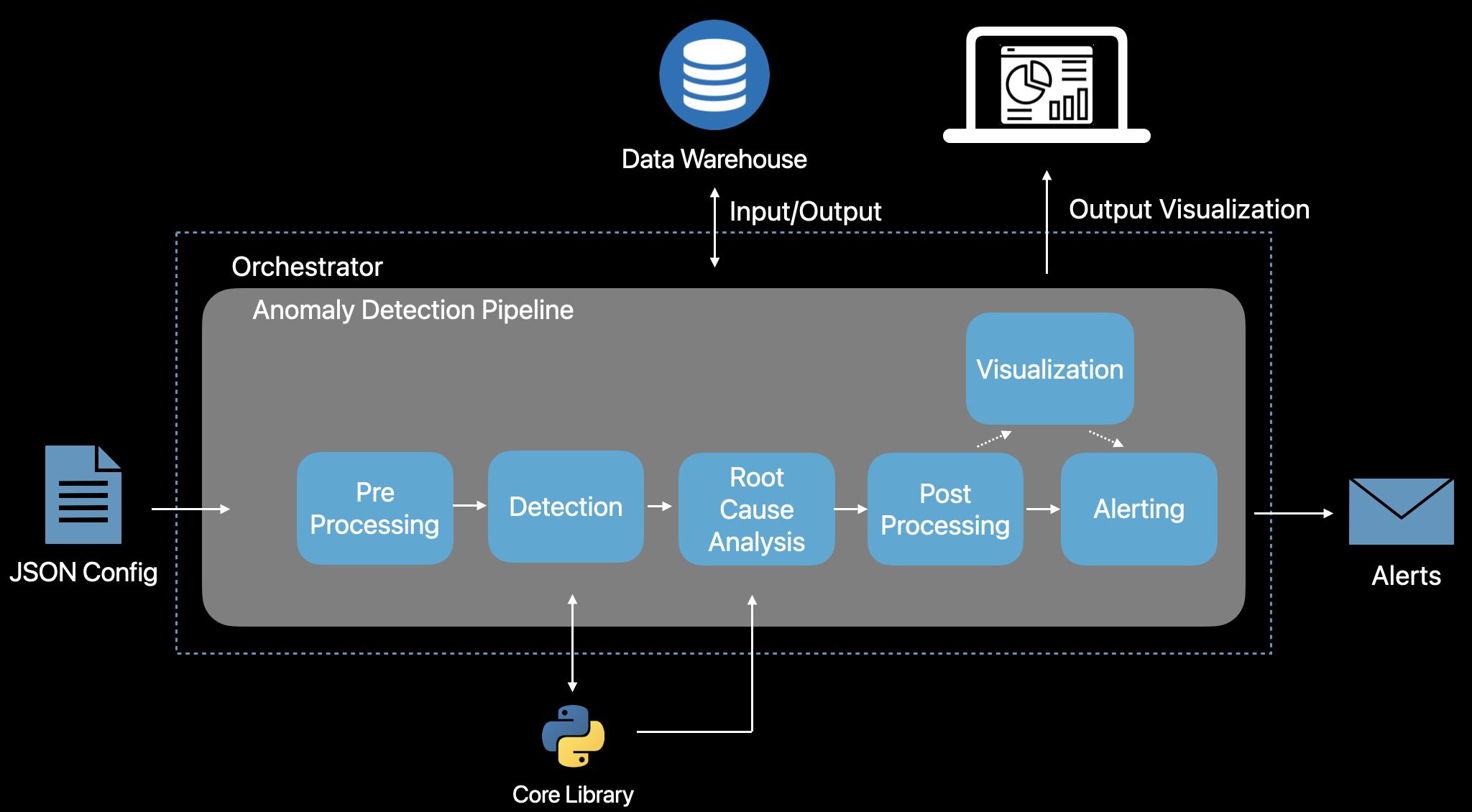} 
    \caption{Anomaly Detection Orchestrator Architecture}
    \label{fig:anomaly_pipeline_arch}
\end{figure}

The RADF Orchestrator is a robust framework designed for programmatically authoring, executing, and monitoring anomaly detection pipelines. It simplifies the creation and execution of these pipelines through a user-friendly configuration file, which utilizes the full range of algorithms available in the Core library. The configuration file defines various pipeline stages, as shown in Figure~\ref{fig:anomaly_pipeline_arch}, including pre-processing, detection, root cause analysis (RCA), post-processing, visualization, and alerting. The pre-processing stage transforms raw data into a format suitable for the detection stage. The detection stage applies anomaly detection algorithms and forwards identified anomalies to the root cause analysis stage, which provides insights into the underlying reasons or causes of the detected anomalies. The post-processing stage performs additional transformations, such as aggregations and summaries, before passing the data to the visualization stage for continuous monitoring or the alerting stage for sending email notifications about business-critical metrics. Users can customize their pipelines by selecting any combination of these stages, with the exception that the alerting stage requires the detection stage to be executed first. A typical RADF pipeline sequence includes: pre-process, detect, root cause analysis, post-process, visualization and/or alerting. 

Pipelines powered by RADF can be deployed as PySpark \cite{zaharia2016apache} jobs for batch anomaly detection or as PyFlink \cite{pyflink2020} jobs for real-time anomaly detection, demonstrating its versatility. Furthermore, the framework supports deployment in distributed environments, enabling the efficient processing of terabytes of data.

The overall architecture of the orchestrator is illustrated in Figure~\ref{fig:anomaly_pipeline_arch}.

\subsection{mSelect}
\label{subsec:algo_mselect}

RADF mSelect recommends the best performing model for a given use case. Similar to the orchestrator configuration file, users can create a Model Select configuration file to leverage the core library's Model Select module that analyzes the data and iterates over different models to identify and recommend the best performing model. 

The Core Library provides multiple algorithms for anomaly detection. However, end users often face challenges in selecting appropriate models and optimizing parameters due to limited understanding of the models and their mechanisms. The RADF mSelect module resolves this issue by employing a novel approach that analyzes input time series data to recommend the most suitable algorithms and parameters for the specific data type, ensuring improved accuracy and efficiency. 

%
%

Even though time series data can be classified in many different ways (for e.g.  regular, irregular, Autoregressive, Moving average), our investigation revealed that the most effective way to recommend an anomaly detection algorithm is by examining whether the time series exhibits a stable, unstable, or trend pattern. The mSelect algorithm, as outlined in Algorithm~\ref{alg:radf_model_select}, classifies time series data into three categories: Stable, Unstable, and Trend. Once categorized, the most suitable ensemble model and corresponding parameters are applied for further analysis. In Step 1, classification begins by determining if the time series exhibits a trend. This is achieved through a rolling median smoother followed by linear regression on the smoothed data. If the regression results indicate a positive coefficient greater than 0.6 and an absolute slope greater than 0.01, the series is classified as a Trend. For series that do not show a trend, the Augmented Dickey-Fuller (ADF) test is applied to assess stationarity. If the ADF test rejects the null hypothesis, the series is classified as Stable; otherwise, it is classified as Unstable. In Step 2, based on the classification result from Step 1, the best algorithms and parameters are recommended. We identified optimal ensemble models based on 18 anomaly detection algorithms described in Table~\ref{tab:algorithm_table} for each category using the benchmarking dataset mentioned in Section~\ref{subsec:eval_mSelect}. Due to business considerations, we will not delve into the specific ensemble models used in this process. However, the application of these models ensures the algorithm is optimized for anomaly detection across various time series patterns, enhancing the accuracy and effectiveness of the results.


Additionally, most datasets lack labeled data, making it even more difficult to identify the right model and parameters. To develop our approach, we evaluated models across 205 time series and 40,000 data points, including those related to billing, subscriptions, and more. These time-series datasets were labeled with actual anomalies, which were used solely to evaluate the performance of the Model Select module. Importantly, the labeling is not required for the module to determine the optimal model and parameters.

\subsection{Root Cause Analysis}

Root Cause Analysis (RCA) identifies the underlying causes of anomalies observed in a system. By isolating the root causes of detected anomalies, corrective actions can be taken to mitigate their recurrence. Within the RADF framework, target time series refer to the series where anomalies have been detected, while candidate time series are those that might potentially explain or contribute to the anomalies in the target time series. For example, if web\_traffic=All is the target time series representing the web traffic across all countries, candidate time series could include web\_traffic=USA, web\_traffic=UK, etc. The RCA process involves evaluating whether anomalies in candidate time series increase the likelihood of anomalies in the target time series. This causal or correlation relationship is quantified using conditional probabilities, with a link established if the conditional probability of an anomaly in the target time series, given an anomaly in the candidate time series, is higher than the baseline probability of an anomaly in the target time series. This concept is mathematically expressed in Equation~\ref{eq:causal_equation}.

\begin{align}
P(T_{\text{target}} \mid T_{\text{candidate}}) &> P(T_{\text{target}}), \label{eq:causal_equation}
\end{align}

\noindent
\text{where:}
\(T_{\text{target}}\): Target time series.
\(T_{\text{candidate}}\): Candidate time series.
\(P(T_{\text{target}})\): Probability of an anomaly in the target time series.
\(P(T_{\text{target}} \mid T_{\text{candidate}})\): Conditional probability of an anomaly in the target time series given an anomaly in the candidate time series.

In anomaly detection pipelines, RCA helps determine an anomaly occurred by analyzing relationships and correlations between multiple variables or events. Currently, RADF supports four RCA methods: Pearson Correlation Coefficient \cite{harary2024efficientalgorithmssensitivitiespearson}, Dynamic Time Warping \cite{NEURIPS2019_02f063c2}, Euclidean Distance, and Granger Analysis \cite{Amornbunchornvej_2021} but can be easily extended to other algorithms as well.

\begin{figure}[htbp]
    \centering
    \includegraphics[width=0.9\linewidth]{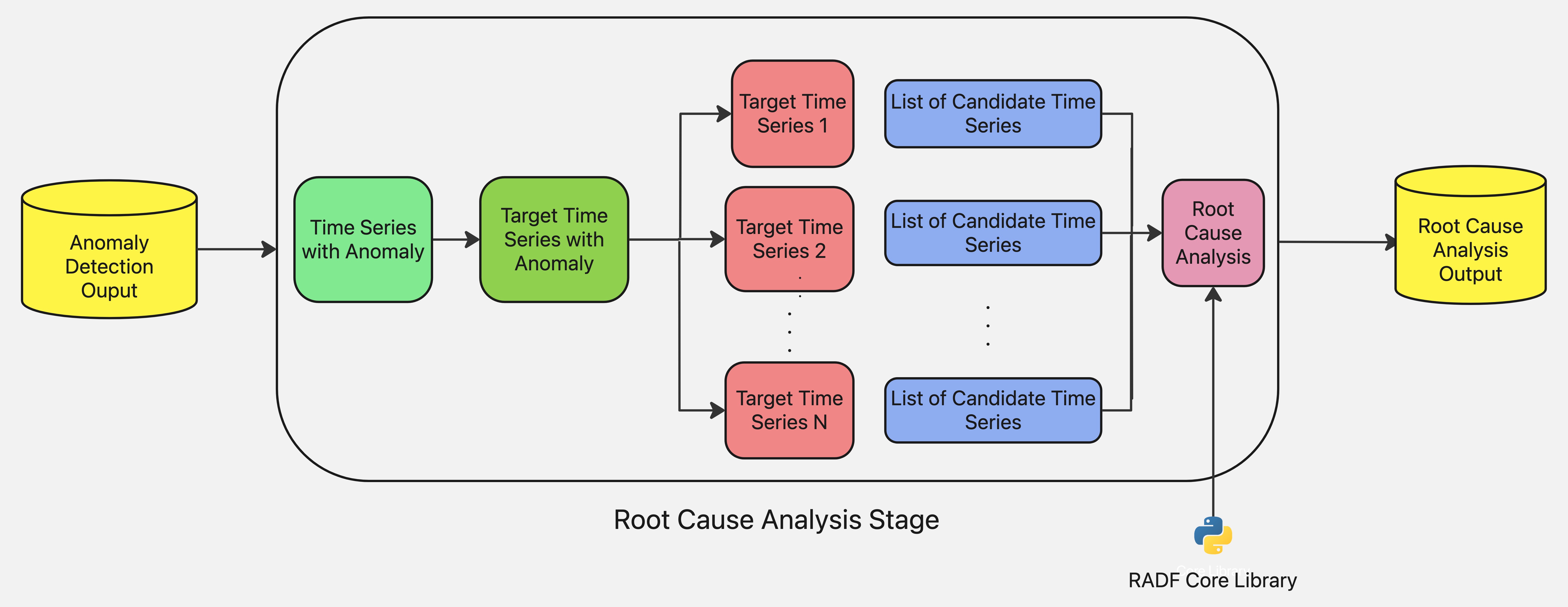} 
    \caption{Workflow for Root Cause Analysis}
    \label{fig:rca_flow}
\end{figure}

After the anomaly detection stage, users can specify the target time series. RADF identifies a set of correlated time series as candidates, either through dimension decomposition (cross-dimensional analysis) or metric correlation (cross-metric analysis). The RCA module then evaluates all pairs of target and candidate time series to uncover causal or correlation relationships using RCA methods. It attributes anomalies in the affected target time series to their corresponding causative candidate time series, as illustrated in the workflow Figure~\ref{fig:rca_flow}.

There are two types of RCA provided by the RADF

\begin{itemize}  
    \item \textbf{Cross-Dimension RCA:} Analyzes interactions between different dimensions of a metric (e.g., region, user group, product category). For each metric, a target dimension is selected, while other dimensions serve as candidates. The framework identifies whether anomalies in the target dimension are influenced by anomalies in the candidate dimensions, helping to pinpoint potential sources of irregularities.  
    \item \textbf{Cross-Metric RCA:} Investigates dependencies between different metrics to uncover anomaly causes. A target metric (e.g., revenue) is examined alongside candidate metrics (e.g., conversion rate, website traffic) to analyze how anomalies in related metrics contribute to deviations in the target. This approach helps in diagnosing underlying issues by understanding metric relationships.  
\end{itemize}  

%
%
%
%


\section{Experiments}
\label{experiments}

In this section, we evaluate RADF using various anomaly detection datasets collected from real-world environments to answer the following questions:

\begin{itemize}
 
        \item What is the overall effectiveness of RADF compared to state-of-the-art methods and frameworks?
	\item Does model selection help improve the effectiveness of RADF?
	\item How does RADF scale for larger datasets?
\end{itemize}

For experimentation, we used the benchmarking algorithms and methods provided by the TSB-UAD \cite{10.14778/3529337.3529354}. TSB-UAD contains 13,766 univariate time series with labeled anomalies spanning different domains with high variability in anomaly types, ratios, and sizes.

For the baseline models, we selected seven unsupervised methods and two deep learning-based semi-supervised models that require anomaly-free training data. These nine models were chosen because they are among the best-performing models on the TSB-UAD \cite{10.14778/3529337.3529354} benchmarking datasets. The deep learning models from TSB-UAD were trained on the initial regions of the time series, with the training ratio set to 30\% for the YAHOO dataset and 10\% for the remaining datasets.

For the initial evaluation, we considered the following strong baselines: Isolation Forest (IForest) \cite{4781136}, which constructs binary trees based on space splitting, where nodes with shorter path lengths to the root are more likely anomalies. We used IForest with a sliding window. IForest used the default 100 base estimators in the tree ensemble. The Local Outlier Factor (LOF) \cite{10.1145/335191.335388} computes the ratio of neighboring density to local density. For LOF, we used 20 as the number of neighbors, following the default settings. Principal Component Analysis (PCA) \cite{10.5555/3086742} projects data onto a lower-dimensional hyperplane, identifying outliers as points with significant distances from this plane. Ten principal components were used for PCA. One-Class Support Vector Machines (OCSVM) \cite{10.5555/3009657.3009740} fit the dataset to define the boundary for normal data. For OCSVM, the upper bound on the fraction of training errors was set to 0.05. Long Short-Term Memory Networks for Anomaly Detection (LSTM-AD) \cite{lstm-malhotra}, Polynomial Approximation (POLY) \cite{poly-zhi}, and Convolutional Neural Networks (CNN) \cite{8581424} detect anomalies by analyzing the deviation between predicted and actual values. LSTM-AD consisted of two LSTM layers with 50 units each, followed by a Dense layer with a single unit, trained with a batch size of 64, up to 50 epochs, and patience of 5. CNNs used three convolutional blocks (filters: 8, 16, 32; kernel size: 2; strides: 1) with max pooling (pool size: 2) and ReLU activation, followed by a Dense layer with 64 units, a Dropout layer with a rate of 0.2, and a Dense output layer. Training used MSE loss, the Adam optimizer, a validation split ratio of 0.15, a batch size of 64, up to 100 epochs, and patience of 5. Discord Aware Matrix Profile (DAMP) \cite{yeh-2016} utilizes the Matrix Profile approach to detect normal patterns by clustering subsequences of the time-series as new data arrives, calculating each point’s effective distance to the normal pattern. This process allows for dynamic updates to the model, making it applicable in both online and offline scenarios. The SAND (Streaming Subsequence Anomaly Detection) algorithm \cite{10.14778/3476311.3476365} is specifically tailored for real-time anomaly detection in streaming time-series data. It works by extracting subsequences of fixed length from the incoming data stream and comparing them to a dynamically updated normal profile. SAND uses efficient similarity measures, such as Euclidean distance or more advanced techniques like Matrix Profiles, to evaluate how closely each new subsequence matches the learned normal behavior.

\subsection{\textbf{Datasets and Evaluation}}

\textbf{Public Datasets:} We evaluated our model on nine publicly available datasets from the TSB-UAD benchmark, covering domains like medical applications and web traffic. These datasets include univariate and multivariate time series, with the latter converted to univariate by labeling each point as normal or anomalous.

The IOPS dataset captures performance indicators reflecting the scale, quality of web services, and machine health status. MGAB \cite{DBLP:data/10/ThillKB20a} consists of Mackey-Glass time series, known for their chaotic behavior and non-trivial anomalies that are challenging for human detection. SensorScope \cite{10.1016/j.peva.2010.08.018} is a collection of environmental data, such as temperature, humidity, and solar radiation, collected from tiered sensor measurement systems. The Yahoo dataset \cite{laptev2015benchmark}, published by Yahoo Labs, consists of real and synthetic time series derived from real production traffic in Yahoo systems. Daphnet \cite{10.1109/TITB.2009.2036165} contains annotated readings from acceleration sensors placed on Parkinson’s disease patients experiencing freezing of gait (FoG) during walking tasks. The GHL dataset \cite{filonov2016multivariateindustrialtimeseries}, also known as the Gasoil Heating Loop Dataset, includes data from three reservoirs, capturing variables such as temperature and level, with anomalies indicating changes in maximum temperature or pump frequency. Genesis \cite{vonBirgelen2018} is a dataset from a portable pick-and-place demonstrator using an air tank for gripping and storage units. OPPORTUNITY (OPP) \cite{5573462} is designed for benchmarking human activity recognition algorithms and includes motion sensor readings collected during typical daily activities. SMD \cite{10.1145/3292500.3330672}, or the Server Machine Dataset, is a 5-week-long dataset from a large Internet company, featuring data from three groups of entities across 28 different machines. These datasets enable rigorous evaluation of anomaly detection across diverse contexts.

For model evaluation, we used AUC and F-measure, which offer a comprehensive assessment of anomaly detection performance. AUC gives a global view of the model’s ability to distinguish between normal and anomalous points, while F-measure provides insights into precision and recall. Additionally, we assessed RADF performance using VUS (Volume under the Surface) measures \cite{10.14778/3551793.3551830}, as discussed in Section~\ref{sec:additional_results}.

\subsection{\textbf{Overall Performance}}

The evaluation of RADF in Tables \ref{tab:performance_comparison} and \ref{tab:vus_performance_comparison} demonstrates its strong performance across diverse datasets. RADF excels in key metrics like AUC and VUS ROC, critical for anomaly detection. Its effectiveness stems from an ensemble approach that integrates multiple algorithms and employs majority voting to identify anomalies, ensuring robustness in complex time-series datasets.

\begin{table*}[ht]
\centering
\scriptsize 
\caption{Performance comparison across AUC and F1 measure metrics for different benchmarking datasets. Bold values indicate the best AUC result for each dataset.}
\label{tab:performance_comparison}
\resizebox{0.9\textwidth}{!}{ 
\begin{tabular}{lcccccccccccccccccccccccccc}
\toprule
\multirow{2}{*}{\textbf{Dataset}} & \multicolumn{2}{c}{\textbf{RADF}} & \multicolumn{2}{c}{\textbf{IFOREST}} & \multicolumn{2}{c}{\textbf{CNN}} & \multicolumn{2}{c}{\textbf{POLY}} & \multicolumn{2}{c}{\textbf{DAMP}} & \multicolumn{2}{c}{\textbf{OCSVM}} & \multicolumn{2}{c}{\textbf{PCA}} & \multicolumn{2}{c}{\textbf{SAND}} & \multicolumn{2}{c}{\textbf{LSTM}} & \multicolumn{2}{c}{\textbf{LOF}} \\
\cmidrule(r){2-3} \cmidrule(r){4-5} \cmidrule(r){6-7} \cmidrule(r){8-9} \cmidrule(r){10-11} \cmidrule(r){12-13} \cmidrule(r){14-15} \cmidrule(r){16-17} \cmidrule(r){18-19} \cmidrule(r){20-21}
                                    & AUC & F   & AUC & F   & AUC & F   & AUC & F   & AUC & F   & AUC & F   & AUC & F   & AUC & F   & AUC & F   & AUC & F   \\
\midrule
YAHOO         & \textbf{0.99} & 0.40 & 0.97 & 0.08 & 0.98 & 0.57 & 0.98 & 0.08 & 0.43 & 0.00 & 0.88 & 0.00 & 0.98 & 0.08 & 0.04 & 0.00 & 0.98 & 0.57 & 0.98 & 0.14 \\
SMD           & 0.90 & 0.15 & 0.85 & 0.35 & 0.72 & 0.07 & 0.92 & 0.15 & 0.43 & 0.00 & 0.75 & 0.37 &  \textbf{0.99} & 0.17 & 0.55 & 0.00 & 0.32 & 0.03 & 0.41 & 0.00 \\
SensorScope   & \textbf{0.85} & 0.01 & 0.65 & 0.03 & 0.48 & 0.00 & 0.79 & 0.10 & 0.68 & 0.00 & 0.46 & 0.00 & 0.50 & 0.09 & 0.36 & 0.00 & 0.38 & 0.00 & 0.69 & 0.01 \\
OPPORTUNITY   & \textbf{0.79} & 0.06 & 0.79 & 0.16 & 0.52 & 0.00 & 0.57 & 0.00 & 0.54 & 0.00 & 0.55 & 0.07 & 0.94 & 0.17 & 0.57 & 0.00 & 0.52 & 0.00 & 0.39 & 0.12 \\
MGAB          & 0.71 & 0.01 & 0.71 & 0.00 & 0.76 & 0.03 & 0.72 & 0.00 & 0.61 & 0.00 & 0.59 & 0.00 & 0.73 & 0.00 & 0.40 & 0.00 & 0.62 & 0.03 & \textbf{0.96} & 0.62 \\
IOPS          & 0.87 & 0.07 & 0.58 & 0.01 & 0.59 & 0.01 & 0.53 & 0.00 & 0.38 & 0.00 & \textbf{0.99} & 0.06 & 0.51 & 0.00 & 0.74 & 0.00 & 0.39 & 0.07 & 0.90 & 0.26 \\
GHL           & \textbf{0.96} & 0.07 & 0.93 & 0.07 & 0.53 & 0.04 & 0.86 & 0.14 & 0.54 & 0.09 & 0.33 & 0.00 & 0.95 & 0.00 & 0.55 & 0.08 & 0.53 & 0.01 & 0.50 & 0.00 \\
Genesis       & \textbf{0.99} & 0.24 & 0.97 & 0.00 & 0.84 & 0.00 & 0.98 & 0.00 & 0.79 & 0.00 & 0.53 & 0.09 & 1.00 & 0.00 & 0.00 & 0.00 & 0.72 & 0.01 & 0.52 & 0.00 \\
Daphnet       & 0.71 & 0.06 & 0.68 & 0.08 & 0.45 & 0.00 & \textbf{0.84} & 0.10 & 0.31 & 0.00 & 0.64 & 0.00 & 0.83 & 0.06 & 0.22 & 0.00 & 0.52 & 0.00 & 0.61 & 0.00 \\
\bottomrule
\end{tabular}
}
\end{table*}

In terms of AUC, RADF consistently outperforms other algorithms, achieving the highest scores in several key datasets. For instance, in the YAHOO dataset, RADF achieves an AUC of 0.99, outperforming competitors such as IFOREST (AUC = 0.97) and CNN (AUC = 0.98). Similarly, RADF demonstrates superior performance in the Genesis (AUC = 0.99) and GHL (AUC = 0.96) datasets, showcasing its ability to adapt to varying data characteristics while maintaining high detection accuracy. RADF also performs competitively in the SMD dataset (AUC = 0.90), despite its inherent complexity, and outperforms, or matches alternative methods, in almost all other datasets evaluated.

However, despite its strong performance in AUC metrics, RADF shows variability in its F1 scores. This behavior is also observed in other algorithms, primarily due to the high class imbalance between anomalous and non-anomalous data points in the datasets \cite{paparrizos2022tsb}.  While it achieves moderate F1 scores in some datasets, such as YAHOO (0.40) and Genesis (0.24), its performance is less consistent in others. For example, in the SMD dataset, RADF achieves an F1 score of only 0.15, suggesting challenges in balancing precision and recall in datasets with high noise levels or imbalanced classes.

Comparatively, RADF often outperforms state-of-the-art algorithms such as CNN, POLY, and IFOREST in most metrics and datasets. For instance, in the YAHOO dataset, RADF achieves the highest AUC score of 0.99, outperforming CNN (AUC = 0.98) and POLY (AUC = 0.98), while maintaining competitive performance F1 scores. In the SMD dataset, RADF matches or exceeds the performance of these methods in AUC (0.90), even though its F1 score is lower. Additionally, RADF demonstrates its adaptability by performing well in highly structured datasets, such as YAHOO and SMD, as well as in more complex datasets, such as Genesis and GHL.

Overall, RADF proves to be a highly effective anomaly detection framework, excelling in AUC metrics, which makes it ideal for applications requiring accurate detection across diverse datasets. Its ability to handle large-scale, complex time-series data highlights its scalability and versatility. However, the variability in F1 scores suggests room for improvement, especially in imbalanced, noisy environments or with high false positive rates. Future work could focus on refining RADF's precision-recall balance. Despite these challenges, RADF’s strong performance across most datasets positions it as a reliable solution for real-world anomaly detection tasks.

\subsection{\textbf{Evaluation of mSelect}}
\label{subsec:eval_mSelect}

The RADF mSelect algorithm was benchmarked using 11 internal datasets, containing approximately 40,000 data points distributed across 205 time series. Each data point is labeled by human evaluators as either an anomaly or normal. The time series data were categorized into three groups: stable, unstable, and trend. Performance was measured using precision, recall, and F1 score. Specifically, the dataset consisted of 124 stable, 72 unstable, and 9 trend time series. Overall, RADF mSelect achieved a precision of 0.978, recall of 0.971, and F1 score of 0.972. The distribution of F1 scores is illustrated in Figure~\ref{fig:mselect_f1score}.

\begin{table}[t]
\captionsetup{font=small} 
\centering
\caption{RADF mSelect Performance metrics}
\label{tab:mselect_results}
\begin{tabular}{lccc} 
\toprule
\textbf{Time Series Type} & \textbf{Precision} & \textbf{Recall} & \textbf{F1 Score} \\ 
\midrule
All         & 0.978 & 0.971 & 0.972 \\ 
Stable      & 0.974 & 0.984 & 0.977 \\ 
Unstable    & 0.990 & 0.982 & 0.986 \\ 
Trend       & 0.947 & 0.695 & 0.798 \\ 
\bottomrule
\end{tabular}%
\end{table}

\begin{figure}[htbp]
    \centering
    \includegraphics[width=0.9\linewidth]{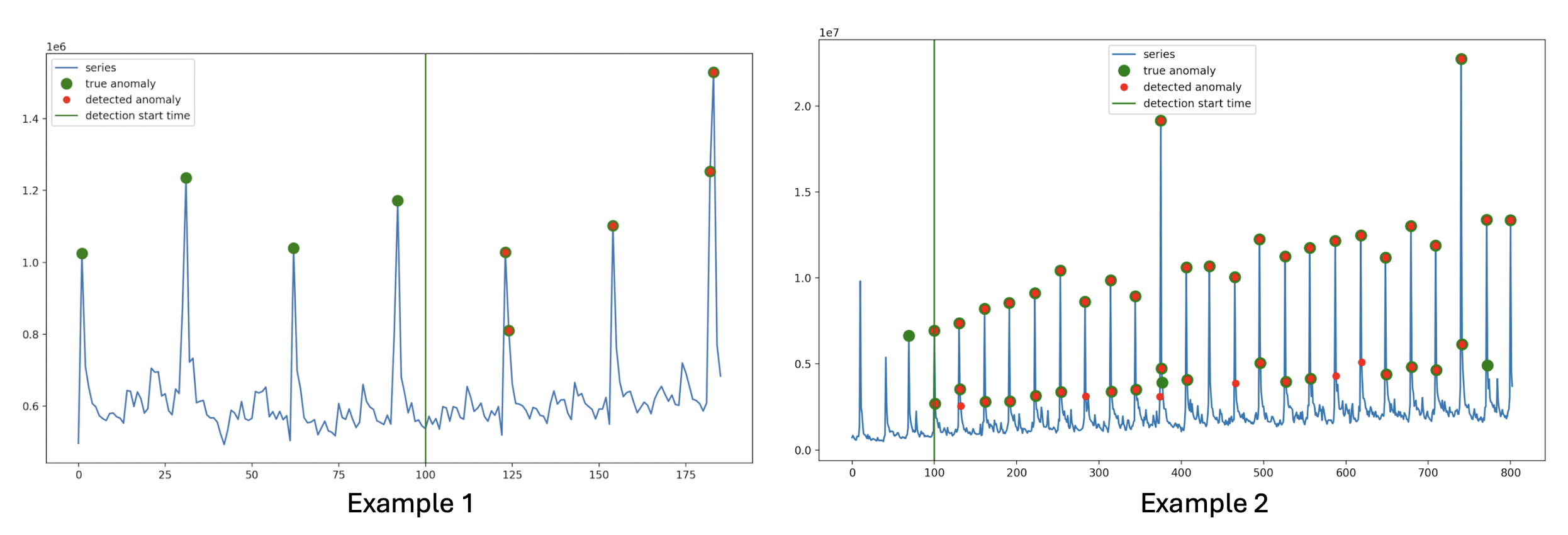} 
    \caption{mSelect Stable Time Series Examples}
    \label{fig:stable_example}
\end{figure}

\textbf{Stable Time series:} For stable time series, RADF mSelect achieved a precision of 0.974, recall of 0.984, and F1 score of 0.977, as shown in Table~\ref{tab:mselect_results}. Examples of its performance on stable time series are depicted in Figure~\ref{fig:stable_example}.

The green vertical lines indicate the start of anomaly detection. In the first example, mSelect correctly identified all five anomalies, while in the second example, it detected all true anomalies except two. These results highlight the robustness of RADF mSelect in detecting anomalies within stable time series.
 
\textbf{Unstable Time series:} In the case of unstable time series, RADF mSelect demonstrated exceptional performance, achieving a precision of 0.990, recall of 0.982, and F1 score of 0.986. These results emphasize the algorithm's strength in handling volatile and unpredictable data, as reflected in Table~\ref{tab:mselect_results}.

\begin{figure}[htbp]
    \centering
    \includegraphics[width=0.9\linewidth]{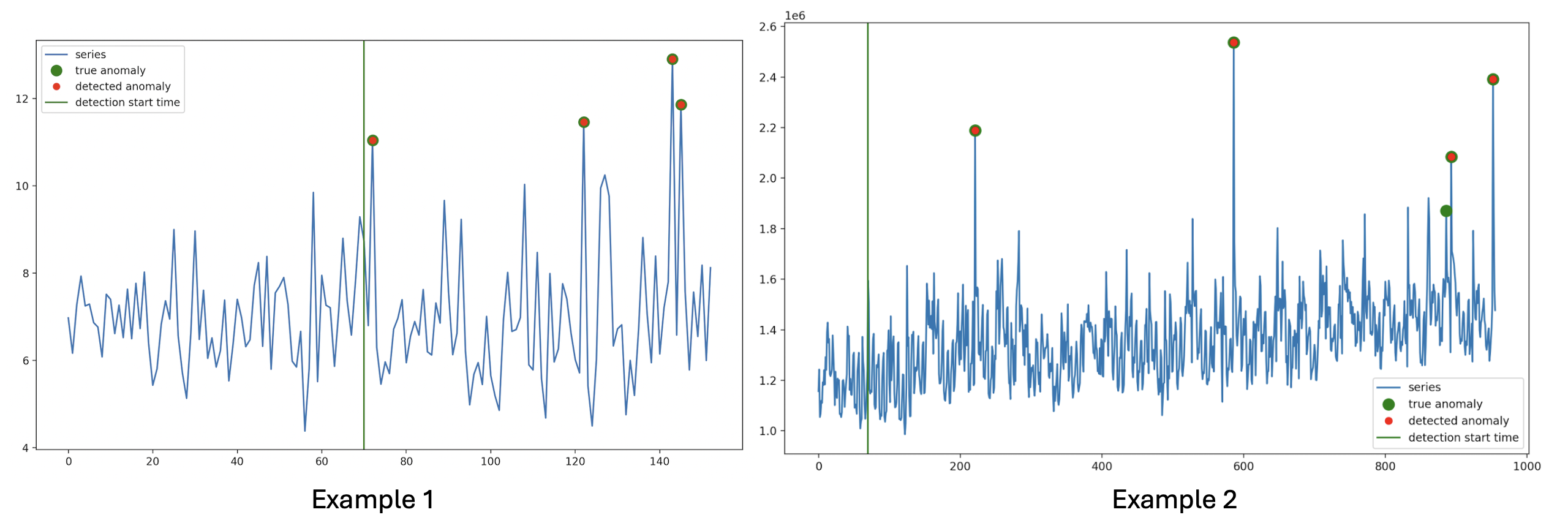} 
    \caption{mSelect Unstable Time Series Examples}
    \label{fig:unstable_example}
\end{figure}

Figure~\ref{fig:unstable_example} showcases examples of unstable time series. RADF mSelect successfully detected all anomalies in most cases, with one false negative and zero false positives, further underlining its accuracy in this category.
 
\textbf{Trend Time series:} RADF mSelect's performance declined for trend time series. It achieved a precision of 0.947, but the recall dropped to 0.695, resulting in an F1 score of 0.798. These results suggest that distinguishing between anomalies and natural trends remains challenging for the algorithm, as indicated in Table~\ref{tab:mselect_results}.

\begin{figure}[htbp]
    \centering
    \includegraphics[width=0.9\linewidth]{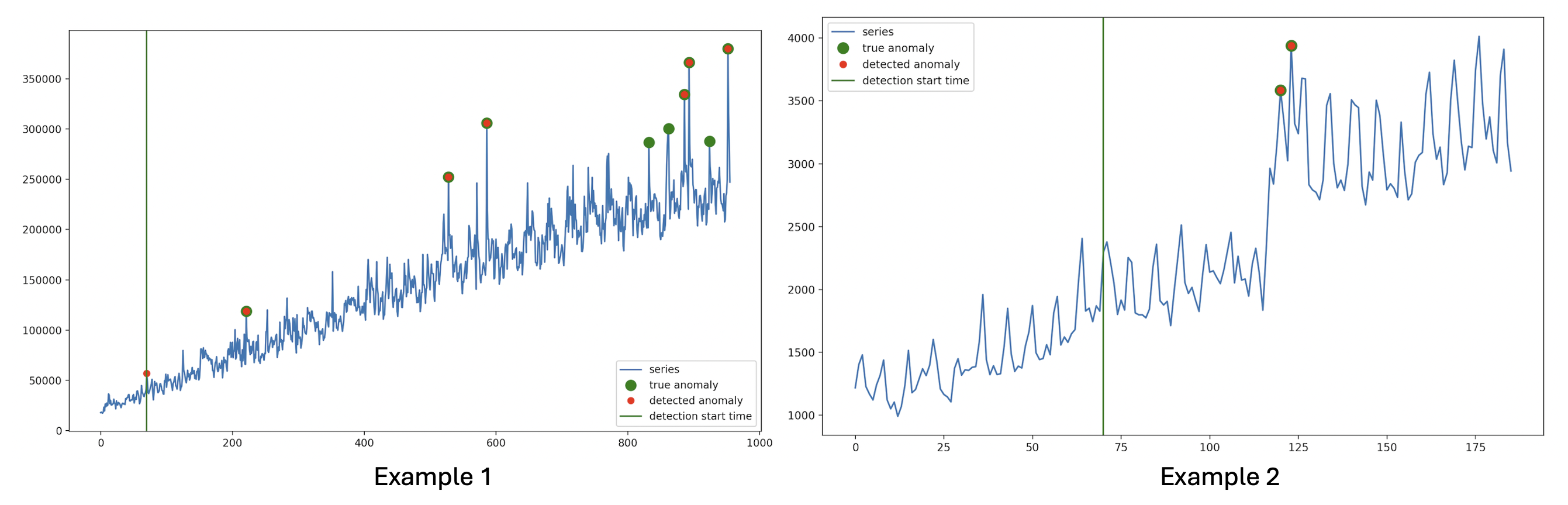} 
    \caption{mSelect Trend Time Series Examples}
    \label{fig:trend_example}
\end{figure}

Figure~\ref{fig:trend_example} presents examples of trend-based time series. In one case, RADF mSelect correctly detected six true anomalies but missed three, resulting in false negatives. In another case, RADF successfully identified both true anomalies without any errors. These results underscore potential areas for improvement in handling trend-based time series. One approach we are exploring is to de-trend the time series before applying RADF.

This analysis provides a comprehensive evaluation of RADF mSelect's performance across different time series categories, offering insights into its strengths and limitations. 

\subsection{\textbf{Deployment}}

RADF has been in production for over three years, supporting more than 30 use cases to detect anomalies at scale across diverse data domains. It has demonstrated robust performance in both batch and streaming pipelines. In batch processing, RADF monitors thousands of internal business KPIs, detecting and alerting on anomalies. In particular, it excels with subscription, commerce and finance datasets where variations are more pronounced. In near-real-time use cases, RADF handles anomaly detection for hundreds of thousands of data points per second, while preserving prediction quality. Modules like mSelect have been instrumental in scaling anomaly detection across thousands of time series datasets. In production, RADF has demonstrated strong performance for both stable and unstable time series without requiring manual intervention. We observed that for time series that display a mixture of trends, manual intervention is needed to fine-tune parameters to achieve optimal results. Our research is currently focussed in addressing this shortcoming.


\section{Conclusion}

The Reasoning-based Anomaly Detection Framework (RADF) introduces a comprehensive, scalable, and real-time approach to anomaly detection across diverse domains and datasets. Its innovative combination of automated model selection, causality-based root cause analysis (RCA), and integration with advanced anomaly detection algorithms sets it apart as a robust solution for handling complex and large-scale time series data. RADF's strong performance across various benchmarking datasets, with competitive precision, recall, and F1 scores, highlights its adaptability and effectiveness in identifying anomalies in stable, unstable, and trend-based time series.

Furthermore, the framework's ability to provide actionable insights through RCA enhances its practical value, allowing stakeholders to quickly diagnose and address issues. The configuration-driven design simplifies deployment and streamlines workflows, making RADF accessible to users without extensive domain expertise. By addressing challenges in model selection, interpretability, and scalability, RADF establishes itself as a state-of-the-art solution for real-world anomaly detection, supporting critical decision-making in high-stakes environments. Future work could focus on enhancing performance for trend time series and further refining the RCA module to address even more complex scenarios. Additionally, a detailed evaluation of the RCA module will be provided to assess its effectiveness in diverse applications.

\bibliographystyle{plainnat}
\bibliography{RADF}

\begin{thebibliography}{66}
\providecommand{\natexlab}[1]{#1}
\providecommand{\url}[1]{\texttt{#1}}
\expandafter\ifx\csname urlstyle\endcsname\relax
  \providecommand{\doi}[1]{doi: #1}\else
  \providecommand{\doi}{doi: \begingroup \urlstyle{rm}\Url}\fi

\bibitem[Aggarwal(2016)]{10.5555/3086742}
Charu~C. Aggarwal.
\newblock \emph{Outlier Analysis}.
\newblock Springer Publishing Company, Incorporated, 2nd edition, 2016.
\newblock ISBN 3319475770.

\bibitem[Ahmed et~al.(2016)Ahmed, Naser~Mahmood, and
  Hu]{10.1016/j.jnca.2015.11.016}
Mohiuddin Ahmed, Abdun Naser~Mahmood, and Jiankun Hu.
\newblock A survey of network anomaly detection techniques.
\newblock \emph{J. Netw. Comput. Appl.}, 60\penalty0 (C):\penalty0 19–31,
  January 2016.
\newblock ISSN 1084-8045.
\newblock \doi{10.1016/j.jnca.2015.11.016}.
\newblock URL \url{https://doi.org/10.1016/j.jnca.2015.11.016}.

\bibitem[Akcay et~al.(2022)Akcay, Ameln, Vaidya, Lakshmanan, Ahuja, and
  Genc]{akcay2022anomalib}
Samet Akcay, Dick Ameln, Ashwin Vaidya, Barath Lakshmanan, Nilesh Ahuja, and
  Utku Genc.
\newblock Anomalib: A deep learning library for anomaly detection.
\newblock In \emph{2022 IEEE International Conference on Image Processing
  (ICIP)}, pages 1706--1710. IEEE, 2022.

\bibitem[Amornbunchornvej et~al.(2021)Amornbunchornvej, Zheleva, and
  Berger-Wolf]{Amornbunchornvej_2021}
Chainarong Amornbunchornvej, Elena Zheleva, and Tanya Berger-Wolf.
\newblock Variable-lag granger causality and transfer entropy for time series
  analysis.
\newblock \emph{ACM Transactions on Knowledge Discovery from Data}, 15\penalty0
  (4):\penalty0 1–30, May 2021.
\newblock ISSN 1556-472X.
\newblock \doi{10.1145/3441452}.
\newblock URL \url{http://dx.doi.org/10.1145/3441452}.

\bibitem[B\"{a}chlin et~al.(2010)B\"{a}chlin, Plotnik, Roggen, Maidan,
  Hausdorff, Giladi, and Tr\"{o}ster]{10.1109/TITB.2009.2036165}
Marc B\"{a}chlin, Meir Plotnik, Daniel Roggen, Inbal Maidan, Jeffrey~M.
  Hausdorff, Nir Giladi, and Gerhard Tr\"{o}ster.
\newblock Wearable assistant for parkinson's disease patients with the freezing
  of gait symptom.
\newblock \emph{Trans. Info. Tech. Biomed.}, 14\penalty0 (2):\penalty0
  436–446, March 2010.
\newblock ISSN 1089-7771.
\newblock \doi{10.1109/TITB.2009.2036165}.
\newblock URL \url{https://doi.org/10.1109/TITB.2009.2036165}.

\bibitem[Bergstra and Bengio(2012)]{bergstra2012random}
James Bergstra and Yoshua Bengio.
\newblock Random search for hyper-parameter optimization.
\newblock \emph{Journal of machine learning research}, 13\penalty0 (2), 2012.

\bibitem[Boniol et~al.(2021)Boniol, Paparrizos, Palpanas, and
  Franklin]{10.14778/3476311.3476365}
Paul Boniol, John Paparrizos, Themis Palpanas, and Michael~J. Franklin.
\newblock Sand in action: subsequence anomaly detection for streams.
\newblock \emph{Proc. VLDB Endow.}, 14\penalty0 (12):\penalty0 2867–2870,
  July 2021.
\newblock ISSN 2150-8097.
\newblock \doi{10.14778/3476311.3476365}.
\newblock URL \url{https://doi.org/10.14778/3476311.3476365}.

\bibitem[Breunig et~al.(2000)Breunig, Kriegel, Ng, and
  Sander]{10.1145/335191.335388}
Markus~M. Breunig, Hans-Peter Kriegel, Raymond~T. Ng, and J\"{o}rg Sander.
\newblock Lof: identifying density-based local outliers.
\newblock \emph{SIGMOD Rec.}, 29\penalty0 (2):\penalty0 93–104, May 2000.
\newblock ISSN 0163-5808.
\newblock \doi{10.1145/335191.335388}.
\newblock URL \url{https://doi.org/10.1145/335191.335388}.

\bibitem[Cai et~al.(2019)Cai, Xu, Yi, Huang, and
  Rajasekaran]{NEURIPS2019_02f063c2}
Xingyu Cai, Tingyang Xu, Jinfeng Yi, Junzhou Huang, and Sanguthevar
  Rajasekaran.
\newblock Dtwnet: a dynamic time warping network.
\newblock In H.~Wallach, H.~Larochelle, A.~Beygelzimer, F.~d\textquotesingle
  Alch\'{e}-Buc, E.~Fox, and R.~Garnett, editors, \emph{Advances in Neural
  Information Processing Systems}, volume~32. Curran Associates, Inc., 2019.
\newblock URL
  \url{https://proceedings.neurips.cc/paper_files/paper/2019/file/02f063c236c7eef66324b432b748d15d-Paper.pdf}.

\bibitem[Chandola et~al.(2009)Chandola, Banerjee, and
  Kumar]{10.1145/1541880.1541882}
Varun Chandola, Arindam Banerjee, and Vipin Kumar.
\newblock Anomaly detection: A survey.
\newblock \emph{ACM Comput. Surv.}, 41\penalty0 (3), July 2009.
\newblock ISSN 0360-0300.
\newblock \doi{10.1145/1541880.1541882}.
\newblock URL \url{https://doi.org/10.1145/1541880.1541882}.

\bibitem[Dau et~al.(2019)Dau, Bagnall, Kamgar, Yeh, Zhu, Gharghabi,
  Ratanamahatana, and Keogh]{dau2019ucrtimeseriesarchive}
Hoang~Anh Dau, Anthony Bagnall, Kaveh Kamgar, Chin-Chia~Michael Yeh, Yan Zhu,
  Shaghayegh Gharghabi, Chotirat~Ann Ratanamahatana, and Eamonn Keogh.
\newblock The ucr time series archive, 2019.
\newblock URL \url{https://arxiv.org/abs/1810.07758}.

\bibitem[Eichler(2013)]{eichler2013causal}
Michael Eichler.
\newblock Causal inference with multiple time series: principles and problems.
\newblock \emph{Philosophical Transactions of the Royal Society A:
  Mathematical, Physical and Engineering Sciences}, 371\penalty0
  (1997):\penalty0 20110613, 2013.

\bibitem[Filonov et~al.(2016)Filonov, Lavrentyev, and
  Vorontsov]{filonov2016multivariateindustrialtimeseries}
Pavel Filonov, Andrey Lavrentyev, and Artem Vorontsov.
\newblock Multivariate industrial time series with cyber-attack simulation:
  Fault detection using an lstm-based predictive data model, 2016.

\bibitem[Geweke(1982)]{geweke1982measurement}
John Geweke.
\newblock Measurement of linear dependence and feedback between multiple time
  series.
\newblock \emph{Journal of the American statistical association}, 77\penalty0
  (378):\penalty0 304--313, 1982.

\bibitem[Granger(1969)]{granger1969investigating}
Clive~WJ Granger.
\newblock Investigating causal relations by econometric models and
  cross-spectral methods.
\newblock \emph{Econometrica: journal of the Econometric Society}, pages
  424--438, 1969.

\bibitem[Han et~al.(2022)Han, Hu, Huang, Jiang, and Zhao]{han2022adbench}
Songqiao Han, Xiyang Hu, Hailiang Huang, Mingqi Jiang, and Yue Zhao.
\newblock Adbench: Anomaly detection benchmark.
\newblock In \emph{Neural Information Processing Systems (NeurIPS)}, 2022.

\bibitem[Harary(2024)]{harary2024efficientalgorithmssensitivitiespearson}
Marc Harary.
\newblock Efficient algorithms for the sensitivities of the pearson correlation
  coefficient and its statistical significance to online data, 2024.
\newblock URL \url{https://arxiv.org/abs/2405.14686}.

\bibitem[Hochenbaum et~al.(2017)Hochenbaum, Vallis, and
  Kejariwal]{hochenbaum2017automatic}
Jordan Hochenbaum, Owen~S. Vallis, and Arun Kejariwal.
\newblock Automatic anomaly detection in the cloud via statistical learning,
  2017.

\bibitem[Hyv{\"a}rinen et~al.(2010)Hyv{\"a}rinen, Zhang, Shimizu, and
  Hoyer]{hyvarinen2010estimation}
Aapo Hyv{\"a}rinen, Kun Zhang, Shohei Shimizu, and Patrik~O Hoyer.
\newblock Estimation of a structural vector autoregression model using
  non-gaussianity.
\newblock \emph{Journal of Machine Learning Research}, 11\penalty0 (5), 2010.

\bibitem[Jiang et~al.(2023)Jiang, Liu, and Xiong]{jiang2023graph}
Wenzhao Jiang, Hao Liu, and Hui Xiong.
\newblock When graph neural network meets causality: Opportunities,
  methodologies and an outlook.
\newblock \emph{arXiv preprint arXiv:2312.12477}, 2023.

\bibitem[Jie et~al.(2024)Jie, Zhou, Su, Zhou, Yuan, Bu, and
  Wang]{10.1145/3589335.3651492}
Xin Jie, Xixi Zhou, Chanfei Su, Zijun Zhou, Yuqing Yuan, Jiajun Bu, and
  Haishuai Wang.
\newblock Disentangled anomaly detection for multivariate time series.
\newblock In \emph{Companion Proceedings of the ACM Web Conference 2024}, WWW
  '24, page 931–934, New York, NY, USA, 2024. Association for Computing
  Machinery.
\newblock ISBN 9798400701726.
\newblock \doi{10.1145/3589335.3651492}.
\newblock URL \url{https://doi.org/10.1145/3589335.3651492}.

\bibitem[Killick et~al.(2012)Killick, Fearnhead, and Eckley]{Killick_2012}
R.~Killick, P.~Fearnhead, and I.~A. Eckley.
\newblock Optimal detection of changepoints with a linear computational cost.
\newblock \emph{Journal of the American Statistical Association}, 107\penalty0
  (500):\penalty0 1590–1598, October 2012.
\newblock ISSN 1537-274X.
\newblock \doi{10.1080/01621459.2012.737745}.
\newblock URL \url{http://dx.doi.org/10.1080/01621459.2012.737745}.

\bibitem[Klementiev et~al.(2009)Klementiev, Roth, Small, and
  Titov]{klementiev2009unsupervised}
Alexandre Klementiev, Dan Roth, Kevin Small, and Ivan Titov.
\newblock Unsupervised rank aggregation with domain-specific expertise.
\newblock In \emph{Twenty-First International Joint Conference on Artificial
  Intelligence}, 2009.

\bibitem[Lai et~al.(2021)Lai, Zha, Wang, Xu, Zhao, Kumar, Chen, Zumkhawaka,
  Wan, Martinez, and
  Hu]{Lai_Zha_Wang_Xu_Zhao_Kumar_Chen_Zumkhawaka_Wan_Martinez_Hu_2021}
Kwei-Herng Lai, Daochen Zha, Guanchu Wang, Junjie Xu, Yue Zhao, Devesh Kumar,
  Yile Chen, Purav Zumkhawaka, Minyang Wan, Diego Martinez, and Xia Hu.
\newblock Tods: An automated time series outlier detection system.
\newblock \emph{Proceedings of the AAAI Conference on Artificial Intelligence},
  35\penalty0 (18):\penalty0 16060--16062, May 2021.

\bibitem[Laptev et~al.(2015{\natexlab{a}})Laptev, Amizadeh, and
  Billawala]{laptev2015benchmark}
Nikolay Laptev, Saeed Amizadeh, and Youssef Billawala.
\newblock A benchmark dataset for time series anomaly detection.
\newblock \emph{von Yahoo Research}, 2015{\natexlab{a}}.

\bibitem[Laptev et~al.(2015{\natexlab{b}})Laptev, Amizadeh, and
  Flint]{laptev2015generic}
Nikolay Laptev, Saeed Amizadeh, and Ian Flint.
\newblock Generic and scalable framework for automated time-series anomaly
  detection.
\newblock In \emph{Proceedings of the 21th ACM SIGKDD International Conference
  on Knowledge Discovery and Data Mining}, pages 1939--1947. ACM,
  2015{\natexlab{b}}.

\bibitem[Lazarevic and Kumar(2005)]{lazarevic2005feature}
Aleksandar Lazarevic and Vipin Kumar.
\newblock Feature bagging for outlier detection.
\newblock In \emph{Proceedings of the eleventh ACM SIGKDD international
  conference on Knowledge discovery in data mining}, pages 157--166, 2005.

\bibitem[Li et~al.(2006)Li, Ma, and Zhou]{poly-zhi}
Zhi Li, Hong Ma, and Yongdao Zhou.
\newblock A unifying method for outlier and change detection from data streams.
\newblock volume~1, pages 580 -- 585, 12 2006.
\newblock \doi{10.1109/ICCIAS.2006.294202}.

\bibitem[Li et~al.(2022)Li, Zhao, Geng, Zhao, Wang, Chen, Jiang, Vaidya, Su,
  and Pei]{9844802}
Zhihan Li, Youjian Zhao, Yitong Geng, Zhanxiang Zhao, Hanzhang Wang, Wenxiao
  Chen, Huai Jiang, Amber Vaidya, Liangfei Su, and Dan Pei.
\newblock Situation-aware multivariate time series anomaly detection through
  active learning and contrast vae-based models in large distributed systems.
\newblock \emph{IEEE Journal on Selected Areas in Communications}, 40\penalty0
  (9):\penalty0 2746--2765, 2022.
\newblock \doi{10.1109/JSAC.2022.3191341}.

\bibitem[Liu et~al.(2015)Liu, Zhao, Xu, Sun, Pei, Luo, Jing, and
  Feng]{10.1145/2815675.2815679}
Dapeng Liu, Youjian Zhao, Haowen Xu, Yongqian Sun, Dan Pei, Jiao Luo, Xiaowei
  Jing, and Mei Feng.
\newblock Opprentice: Towards practical and automatic anomaly detection through
  machine learning.
\newblock In \emph{Proceedings of the 2015 Internet Measurement Conference},
  IMC '15, page 211–224, New York, NY, USA, 2015. Association for Computing
  Machinery.
\newblock ISBN 9781450338486.
\newblock \doi{10.1145/2815675.2815679}.
\newblock URL \url{https://doi.org/10.1145/2815675.2815679}.

\bibitem[Liu et~al.(2008)Liu, Ting, and Zhou]{4781136}
Fei~Tony Liu, Kai~Ming Ting, and Zhi-Hua Zhou.
\newblock Isolation forest.
\newblock In \emph{2008 Eighth IEEE International Conference on Data Mining},
  pages 413--422, 2008.
\newblock \doi{10.1109/ICDM.2008.17}.

\bibitem[Malhotra et~al.(2015)Malhotra, Vig, Shroff, and
  Agarwal]{lstm-malhotra}
Pankaj Malhotra, Lovekesh Vig, Gautam Shroff, and Puneet Agarwal.
\newblock Long short term memory networks for anomaly detection in time series.
\newblock 04 2015.

\bibitem[Malinsky and Spirtes(2018)]{malinsky2018causal}
Daniel Malinsky and Peter Spirtes.
\newblock Causal structure learning from multivariate time series in settings
  with unmeasured confounding.
\newblock In \emph{Proceedings of 2018 ACM SIGKDD workshop on causal
  discovery}, pages 23--47. PMLR, 2018.

\bibitem[Mensi and Bicego(2021)]{10.1016/j.patcog.2021.108115}
Antonella Mensi and Manuele Bicego.
\newblock Enhanced anomaly scores for isolation forests.
\newblock \emph{Pattern Recogn.}, 120\penalty0 (C), December 2021.
\newblock ISSN 0031-3203.
\newblock \doi{10.1016/j.patcog.2021.108115}.
\newblock URL \url{https://doi.org/10.1016/j.patcog.2021.108115}.

\bibitem[Munir et~al.(2019)Munir, Siddiqui, Dengel, and Ahmed]{8581424}
Mohsin Munir, Shoaib~Ahmed Siddiqui, Andreas Dengel, and Sheraz Ahmed.
\newblock Deepant: A deep learning approach for unsupervised anomaly detection
  in time series.
\newblock \emph{IEEE Access}, 7:\penalty0 1991--2005, 2019.
\newblock \doi{10.1109/ACCESS.2018.2886457}.

\bibitem[Nauta et~al.(2019)Nauta, Bucur, and Seifert]{nauta2019causal}
Meike Nauta, Doina Bucur, and Christin Seifert.
\newblock Causal discovery with attention-based convolutional neural networks.
\newblock \emph{Machine Learning and Knowledge Extraction}, 1\penalty0
  (1):\penalty0 19, 2019.

\bibitem[Navarro et~al.(2023)Navarro, Huet, and Rossi]{navarro2023meta}
Jose~Manuel Navarro, Alexis Huet, and Dario Rossi.
\newblock Meta-learning for fast model recommendation in unsupervised
  multivariate time series anomaly detection.
\newblock In \emph{International Conference on Automated Machine Learning},
  pages 24--1. PMLR, 2023.

\bibitem[Pamfil et~al.(2020)Pamfil, Sriwattanaworachai, Desai, Pilgerstorfer,
  Georgatzis, Beaumont, and Aragam]{pamfil2020dynotears}
Roxana Pamfil, Nisara Sriwattanaworachai, Shaan Desai, Philip Pilgerstorfer,
  Konstantinos Georgatzis, Paul Beaumont, and Bryon Aragam.
\newblock Dynotears: Structure learning from time-series data.
\newblock In \emph{International Conference on Artificial Intelligence and
  Statistics}, pages 1595--1605. Pmlr, 2020.

\bibitem[Paparrizos et~al.(2022{\natexlab{a}})Paparrizos, Boniol, Palpanas,
  Tsay, Elmore, and Franklin]{10.14778/3551793.3551830}
John Paparrizos, Paul Boniol, Themis Palpanas, Ruey~S. Tsay, Aaron Elmore, and
  Michael~J. Franklin.
\newblock Volume under the surface: a new accuracy evaluation measure for
  time-series anomaly detection.
\newblock \emph{Proc. VLDB Endow.}, 15\penalty0 (11):\penalty0 2774–2787,
  July 2022{\natexlab{a}}.
\newblock ISSN 2150-8097.
\newblock \doi{10.14778/3551793.3551830}.
\newblock URL \url{https://doi.org/10.14778/3551793.3551830}.

\bibitem[Paparrizos et~al.(2022{\natexlab{b}})Paparrizos, Kang, Boniol, Tsay,
  Palpanas, and Franklin]{10.14778/3529337.3529354}
John Paparrizos, Yuhao Kang, Paul Boniol, Ruey~S. Tsay, Themis Palpanas, and
  Michael~J. Franklin.
\newblock Tsb-uad: an end-to-end benchmark suite for univariate time-series
  anomaly detection.
\newblock \emph{Proc. VLDB Endow.}, 15\penalty0 (8):\penalty0 1697–1711,
  April 2022{\natexlab{b}}.
\newblock ISSN 2150-8097.
\newblock \doi{10.14778/3529337.3529354}.
\newblock URL \url{https://doi.org/10.14778/3529337.3529354}.

\bibitem[Paparrizos et~al.(2022{\natexlab{c}})Paparrizos, Kang, Boniol, Tsay,
  Palpanas, and Franklin]{paparrizos2022tsb}
John Paparrizos, Yuhao Kang, Paul Boniol, Ruey~S Tsay, Themis Palpanas, and
  Michael~J Franklin.
\newblock Tsb-uad: an end-to-end benchmark suite for univariate time-series
  anomaly detection.
\newblock \emph{Proceedings of the VLDB Endowment}, 15\penalty0 (8):\penalty0
  1697--1711, 2022{\natexlab{c}}.

\bibitem[Park et~al.(2018)Park, Hoshi, and Kemp]{Park_2018}
Daehyung Park, Yuuna Hoshi, and Charles~C. Kemp.
\newblock A multimodal anomaly detector for robot-assisted feeding using an
  lstm-based variational autoencoder.
\newblock \emph{IEEE Robotics and Automation Letters}, 3\penalty0 (3):\penalty0
  1544–1551, July 2018.
\newblock ISSN 2377-3774.
\newblock \doi{10.1109/lra.2018.2801475}.
\newblock URL \url{http://dx.doi.org/10.1109/LRA.2018.2801475}.

\bibitem[Peters et~al.(2016)Peters, B{\"u}hlmann, and
  Meinshausen]{peters2016causal}
Jonas Peters, Peter B{\"u}hlmann, and Nicolai Meinshausen.
\newblock Causal inference by using invariant prediction: identification and
  confidence intervals.
\newblock \emph{Journal of the Royal Statistical Society Series B: Statistical
  Methodology}, 78\penalty0 (5):\penalty0 947--1012, 2016.

\bibitem[Ragab et~al.(2023)Ragab, Eldele, Tan, Foo, Chen, Wu, Kwoh, and
  Li]{adatime}
Mohamed Ragab, Emadeldeen Eldele, Wee~Ling Tan, Chuan-Sheng Foo, Zhenghua Chen,
  Min Wu, Chee-Keong Kwoh, and Xiaoli Li.
\newblock Adatime: A benchmarking suite for domain adaptation on time series
  data.
\newblock \emph{ACM Trans. Knowl. Discov. Data}, mar 2023.
\newblock ISSN 1556-4681.
\newblock \doi{10.1145/3587937}.
\newblock URL \url{https://doi.org/10.1145/3587937}.

\bibitem[Rayana and Akoglu(2016)]{rayana2016less}
Shebuti Rayana and Leman Akoglu.
\newblock Less is more: Building selective anomaly ensembles.
\newblock \emph{Acm transactions on knowledge discovery from data (tkdd)},
  10\penalty0 (4):\penalty0 1--33, 2016.

\bibitem[Roggen et~al.(2010)Roggen, Calatroni, Rossi, Holleczek, Förster,
  Tröster, Lukowicz, Bannach, Pirkl, Ferscha, Doppler, Holzmann, Kurz, Holl,
  Chavarriaga, Sagha, Bayati, Creatura, and Millàn]{5573462}
Daniel Roggen, Alberto Calatroni, Mirco Rossi, Thomas Holleczek, Kilian
  Förster, Gerhard Tröster, Paul Lukowicz, David Bannach, Gerald Pirkl, Alois
  Ferscha, Jakob Doppler, Clemens Holzmann, Marc Kurz, Gerald Holl, Ricardo
  Chavarriaga, Hesam Sagha, Hamidreza Bayati, Marco Creatura, and José del~R.
  Millàn.
\newblock Collecting complex activity datasets in highly rich networked sensor
  environments.
\newblock In \emph{2010 Seventh International Conference on Networked Sensing
  Systems (INSS)}, pages 233--240, 2010.
\newblock \doi{10.1109/INSS.2010.5573462}.

\bibitem[Runge(2020)]{runge2020discovering}
Jakob Runge.
\newblock Discovering contemporaneous and lagged causal relations in
  autocorrelated nonlinear time series datasets.
\newblock In \emph{Conference on Uncertainty in Artificial Intelligence}, pages
  1388--1397. Pmlr, 2020.

\bibitem[Sch\"{o}lkopf et~al.(1999)Sch\"{o}lkopf, Williamson, Smola,
  Shawe-Taylor, and Platt]{10.5555/3009657.3009740}
Bernhard Sch\"{o}lkopf, Robert Williamson, Alex Smola, John Shawe-Taylor, and
  John Platt.
\newblock Support vector method for novelty detection.
\newblock In \emph{Proceedings of the 12th International Conference on Neural
  Information Processing Systems}, NIPS'99, page 582–588, Cambridge, MA, USA,
  1999. MIT Press.

\bibitem[Schubert et~al.(2012)Schubert, Kriegel, Kr{\"o}ger, and
  Zimek]{schubert2012evaluation}
Erich Schubert, Hans-Peter Kriegel, Peer Kr{\"o}ger, and Arthur Zimek.
\newblock Evaluation of outlier rankings and outlier scores.
\newblock In \emph{Proceedings of the 2012 SIAM International Conference on
  Data Mining}, pages 1047--1058. SIAM, 2012.

\bibitem[Singh and Vanschoren(2022)]{singh2022meta}
Prabhant Singh and Joaquin Vanschoren.
\newblock Meta-learning for unsupervised outlier detection with optimal
  transport.
\newblock \emph{arXiv preprint arXiv:2211.00372}, 2022.

\bibitem[Snoek et~al.(2012)Snoek, Larochelle, and Adams]{snoek2012practical}
Jasper Snoek, Hugo Larochelle, and Ryan~P Adams.
\newblock Practical bayesian optimization of machine learning algorithms.
\newblock \emph{Advances in neural information processing systems}, 25, 2012.

\bibitem[Su et~al.(2019)Su, Zhao, Niu, Liu, Sun, and
  Pei]{10.1145/3292500.3330672}
Ya~Su, Youjian Zhao, Chenhao Niu, Rong Liu, Wei Sun, and Dan Pei.
\newblock Robust anomaly detection for multivariate time series through
  stochastic recurrent neural network.
\newblock In \emph{Proceedings of the 25th ACM SIGKDD International Conference
  on Knowledge Discovery \& Data Mining}, KDD '19, page 2828–2837, New York,
  NY, USA, 2019. Association for Computing Machinery.
\newblock ISBN 9781450362016.
\newblock \doi{10.1145/3292500.3330672}.
\newblock URL \url{https://doi.org/10.1145/3292500.3330672}.

\bibitem[Sun et~al.(2015)Sun, Taylor, and Bollt]{sun2015causal}
Jie Sun, Dane Taylor, and Erik~M Bollt.
\newblock Causal network inference by optimal causation entropy.
\newblock \emph{SIAM Journal on Applied Dynamical Systems}, 14\penalty0
  (1):\penalty0 73--106, 2015.

\bibitem[Sylligardos et~al.(2023)Sylligardos, Boniol, Paparrizos, Trahanias,
  and Palpanas]{boniolMSAD2023}
Emmanouil Sylligardos, Paul Boniol, John Paparrizos, Panos Trahanias, and
  Themis Palpanas.
\newblock Choose wisely: An extensive evaluation of model selection for anomaly
  detection in time series.
\newblock \emph{Proceedings of the VLDB Endowment}, 16\penalty0 (11):\penalty0
  3418--3432, 2023.

\bibitem[Tai et~al.(2020)Tai, Ewen, Tzoumas, et~al.]{pyflink2020}
Tzu-Li~(Gordon) Tai, Stephan Ewen, Kostas Tzoumas, et~al.
\newblock Pyflink: A python api for apache flink.
\newblock In \emph{Proceedings of the 2020 ACM SIGMOD International Conference
  on Management of Data (SIGMOD '20)}, pages 1905--1913. ACM, 2020.
\newblock \doi{10.1145/3318464.3389724}.
\newblock URL \url{https://doi.org/10.1145/3318464.3389724}.

\bibitem[Thill et~al.(2020)Thill, Konen, and
  B{\"{a}}ck]{DBLP:data/10/ThillKB20a}
Markus Thill, Wolfgang Konen, and Thomas B{\"{a}}ck.
\newblock Markusthill/mgab: The mackey-glass anomaly benchmark (version
  v1.0.1).
\newblock \url{https://doi.org/10.5281/zenodo.3762385}, April 2020.
\newblock URL \url{https://doi.org/10.5281/zenodo.3762385}.
\newblock Accessed on YYYY-MM-DD.

\bibitem[von Birgelen and Niggemann(2018)]{vonBirgelen2018}
Alexander von Birgelen and Oliver Niggemann.
\newblock \emph{Anomaly Detection and Localization for Cyber-Physical
  Production Systems with Self-Organizing Maps}, pages 55--71.
\newblock Springer Berlin Heidelberg, Berlin, Heidelberg, 2018.
\newblock ISBN 978-3-662-57805-6.

\bibitem[Wu et~al.(2020)Wu, He, Lin, Su, Cui, Maple, and
  Jarvis]{wu2020developing}
Wentai Wu, Ligang He, Weiwei Lin, Yi~Su, Yuhua Cui, Carsten Maple, and Stephen
  Jarvis.
\newblock Developing an unsupervised real-time anomaly detection scheme for
  time series with multi-seasonality.
\newblock \emph{IEEE Transactions on Knowledge and Data Engineering},
  34\penalty0 (9):\penalty0 4147--4160, 2020.

\bibitem[Yao et~al.(2010)Yao, Sharma, Golubchik, and
  Govindan]{10.1016/j.peva.2010.08.018}
Yuan Yao, Abhishek Sharma, Leana Golubchik, and Ramesh Govindan.
\newblock Online anomaly detection for sensor systems: A simple and efficient
  approach.
\newblock \emph{Perform. Eval.}, 67\penalty0 (11):\penalty0 1059–1075,
  November 2010.
\newblock ISSN 0166-5316.
\newblock \doi{10.1016/j.peva.2010.08.018}.
\newblock URL \url{https://doi.org/10.1016/j.peva.2010.08.018}.

\bibitem[Yeh et~al.(2016)Yeh, Zhu, Ulanova, Begum, Ding, Dau, Silva, Mueen, and
  Keogh]{yeh-2016}
Chin-Chia~Michael Yeh, Yan Zhu, Liudmila Ulanova, Nurjahan Begum, Yifei Ding,
  Anh Dau, Diego Silva, Abdullah Mueen, and Eamonn Keogh.
\newblock Matrix profile i: All pairs similarity joins for time series: A
  unifying view that includes motifs, discords and shapelets.
\newblock pages 1317--1322, 12 2016.
\newblock \doi{10.1109/ICDM.2016.0179}.

\bibitem[Ying et~al.(2020)Ying, Duan, Wang, Wang, Huang, and
  Xu]{ying2020automated}
Yuanxiang Ying, Juanyong Duan, Chunlei Wang, Yujing Wang, Congrui Huang, and
  Bixiong Xu.
\newblock Automated model selection for time-series anomaly detection.
\newblock \emph{arXiv preprint arXiv:2009.04395}, 2020.

\bibitem[Zaharia et~al.(2016)Zaharia, Xin, Wendell, Das, Armbrust, Dave, Meng,
  Rosen, Venkataraman, Franklin, et~al.]{zaharia2016apache}
Matei Zaharia, Reynold~S Xin, Patrick Wendell, Tathagata Das, Michael Armbrust,
  Ankur Dave, Xiangrui Meng, Josh Rosen, Shivaram Venkataraman, Michael~J
  Franklin, et~al.
\newblock Apache spark: a unified engine for big data processing.
\newblock \emph{Communications of the ACM}, 59\penalty0 (11):\penalty0 56--65,
  2016.

\bibitem[Zhang and Chen(2019)]{Zhang2019TimeSA}
Chun~Kai Zhang and Yingyang Chen.
\newblock Time series anomaly detection with variational autoencoders.
\newblock \emph{ArXiv}, abs/1907.01702, 2019.
\newblock URL \url{https://api.semanticscholar.org/CorpusID:195791542}.

\bibitem[Zhao et~al.(2021)Zhao, Rossi, and Akoglu]{zhao2021automatic}
Yue Zhao, Ryan Rossi, and Leman Akoglu.
\newblock Automatic unsupervised outlier model selection.
\newblock \emph{Advances in Neural Information Processing Systems},
  34:\penalty0 4489--4502, 2021.

\bibitem[Zhao et~al.(2022)Zhao, Zhang, and Akoglu]{zhao2022toward}
Yue Zhao, Sean Zhang, and Leman Akoglu.
\newblock Toward unsupervised outlier model selection.
\newblock In \emph{2022 IEEE International Conference on Data Mining (ICDM)},
  pages 773--782. IEEE, 2022.

\bibitem[Zimek et~al.(2014)Zimek, Campello, and Sander]{zimek2014ensembles}
Arthur Zimek, Ricardo~JGB Campello, and J{\"o}rg Sander.
\newblock Ensembles for unsupervised outlier detection: challenges and research
  questions a position paper.
\newblock \emph{Acm Sigkdd Explorations Newsletter}, 15\penalty0 (1):\penalty0
  11--22, 2014.

\end{thebibliography}
\appendix

\section{Algorithm mSelect}
\label{sec:algo_mselect}

Algorithm~\ref{alg:radf_model_select} presents the pseudocode for mSelect. More details can be found in Section~\ref{subsec:algo_mselect}.

\begin{algorithm}[ht]
\caption{RADF mSelect}
\label{alg:radf_model_select}
\KwIn{Time Series $T$}
\KwOut{Classification of $T$ as Stable, Unstable, or Trend, and find best model and parameters for the time series type}

\textbf{Step 1: Classify Time Series as Stable, Unstable, or Trend}
\begin{enumerate}
    \item \textbf{Step 1.1: Identify if $T$ is a trend or not}
    \begin{enumerate}
        \item Apply rolling median smoother on $T$, resulting in smoothed time series $T_{smooth}$.
        \item Apply linear regression on $T_{smooth}$.
        \item Classify $T$ as a \textbf{trend} if coefficient > 0.6 and absolute slope > 0.01.
    \end{enumerate}
    \item \textbf{Step 1.2: Classify Non-Trend Time Series into Stable and Unstable}
    \begin{enumerate}
        \item Perform Augmented Dickey-Fuller test on $T$.
        \item \textbf{If} ADF test rejects the null hypothesis (non-stationarity), classify $T$ as \textbf{Stable}.
        \item \textbf{Else}, classify $T$ as \textbf{Unstable}.
    \end{enumerate}
\end{enumerate}

\textbf{Step 2: Apply Best Model and Parameters based on Time Series Type and Similarity with Benchmarking datasets}
\end{algorithm}

\section{Additional Results}
\label{sec:additional_results}

\subsection{VUS Performance Comparison}
In this section, we provide extended overall performance Results, including detailed discussion on the evaluation of RADF on VUS measures \cite{10.14778/3551793.3551830}, which are parameter-free and threshold-independent. We compared VUS-ROC and VUS-Precision across datasets to provide robust performance insights.

The results for VUS ROC as shown in Table~\ref{tab:vus_performance_comparison} further highlight RADF's robustness. RADF achieves the best VUS ROC scores in datasets such as YAHOO (0.99), SMD (0.95), and Genesis (0.99), indicating its ability to capture multi-threshold performance effectively. These results demonstrate that RADF not only excels in binary classification tasks but also performs reliably under different decision thresholds, which is critical for real-world applications where anomaly detection systems must operate under varying sensitivity requirements.

Similarly, its Precision scores, while strong in datasets like SMD (0.72) and SensorScope (0.68), are comparatively weaker in datasets like MGAB (0.01) and OPPORTUNITY (0.06). This variability indicates that while RADF excels in detecting anomalies, it may struggle in scenarios where minimizing false positives or achieving a balance between detection sensitivity and specificity is critical.

\begin{table*}[ht]
\centering
\scriptsize 
\caption{Performance comparison across VUS(Volume Under the Surface) ROC and Precision metrics for different benchmarking datasets. Bold values indicate the best VUS ROC result for each dataset.}
\label{tab:vus_performance_comparison}
\resizebox{0.9\textwidth}{!}{ 
\begin{tabular}{lcccccccccccccccccccccccccc}
\toprule
\multirow{2}{*}{\textbf{Dataset}} & \multicolumn{2}{c}{\textbf{RADF}} & \multicolumn{2}{c}{\textbf{IFOREST}} & \multicolumn{2}{c}{\textbf{CNN}} & \multicolumn{2}{c}{\textbf{POLY}} & \multicolumn{2}{c}{\textbf{DAMP}} & \multicolumn{2}{c}{\textbf{OCSVM}} & \multicolumn{2}{c}{\textbf{PCA}} & \multicolumn{2}{c}{\textbf{SAND}} & \multicolumn{2}{c}{\textbf{LSTM}} & \multicolumn{2}{c}{\textbf{LOF}} \\
\cmidrule(r){2-3} \cmidrule(r){4-5} \cmidrule(r){6-7} \cmidrule(r){8-9} \cmidrule(r){10-11} \cmidrule(r){12-13} \cmidrule(r){14-15} \cmidrule(r){16-17} \cmidrule(r){18-19} \cmidrule(r){20-21}
                                    & ROC & PR   & ROC & PR   & ROC & PR   & ROC & PR   & ROC & PR   & ROC & PR   & ROC & PR   & ROC & PR   & ROC & PR   & ROC & PR   \\
\midrule
YAHOO         & \textbf{0.99} & 0.40 & 0.89 & 0.53 & 0.74 & 0.15 & 0.98 & 0.65 & 0.43 & 0.01 & 0.95 & 0.17 & 0.98 & 0.60 & 0.04 & 0.01 & 0.78 & 0.21 & 0.98 & 0.44 \\
SMD           & \textbf{0.95} & 0.72 & 0.92 & 0.67 & 0.73 & 0.23 & 0.82 & 0.43 & 0.43 & 0.11 & 0.84 & 0.57 & 0.94 & 0.78 & 0.58 & 0.14 & 0.42 & 0.10 & 0.53 & 0.15 \\
SensorScope   & \textbf{0.85} & 0.68 & 0.64 & 0.35 & 0.71 & 0.39 & 0.78 & 0.57 & 0.68 & 0.36 & 0.44 & 0.32 & 0.49 & 0.31 & 0.36 & 0.20 & 0.38 & 0.23 & 0.69 & 0.43 \\
OPPORTUNITY   & 0.79 & 0.06 & 0.80 & 0.10 & 0.49 & 0.03 & 0.58 & 0.03 & 0.55 & 0.04 & 0.57 & 0.04 & \textbf{0.94} & 0.21 & 0.57 & 0.04 & 0.54 & 0.03 & 0.41 & 0.05 \\
MGAB          & 0.82 & 0.01 & 0.74 & 0.01 & 0.78 & 0.09 & 0.71 & 0.01 & 0.56 & 0.01 & 0.70 & 0.01 & 0.77 & 0.01 & 0.48 & 0.00 & 0.72 & 0.05 & \textbf{0.97} & 0.48 \\
IOPS          & \textbf{0.95} & 0.21 & 0.71 & 0.11 & 0.70 & 0.09 & 0.60 & 0.12 & 0.59 & 0.08 & 1.00 & 0.30 & 0.61 & 0.10 & 0.45 & 0.07 & 0.74 & 0.10 & 0.83 & 0.29 \\
GHL           & \textbf{0.96} & 0.07 & 0.94 & 0.14 & 0.54 & 0.07 & 0.87 & 0.09 & 0.55 & 0.13 & 0.40 & 0.05 & 0.95 & 0.57 & 0.57 & 0.06 & 0.54 & 0.08 & 0.50 & 0.08 \\
Genesis       & \textbf{0.99} & 0.68 & 0.99 & 0.57 & 0.86 & 0.02 & 0.98 & 0.50 & 0.83 & 0.02 & 0.54 & 0.08 & 0.99 & 0.70 & 0.01 & 0.00 & 0.79 & 0.02 & 0.58 & 0.01 \\
Daphnet       & 0.70 & 0.22 & 0.68 & 0.02 & 0.46 & 0.09 & \textbf{0.84} & 0.05 & 0.32 & 0.08 & 0.57 & 0.39 & 0.83 & 0.04 & 0.22 & 0.01 & 0.52 & 0.01 & 0.61 & 0.01 \\
\bottomrule
\end{tabular}
}
\end{table*}

\subsection{Analysis of mSelect results}
\begin{figure}[htbp]
    \centering
    \includegraphics[width=0.9\linewidth]{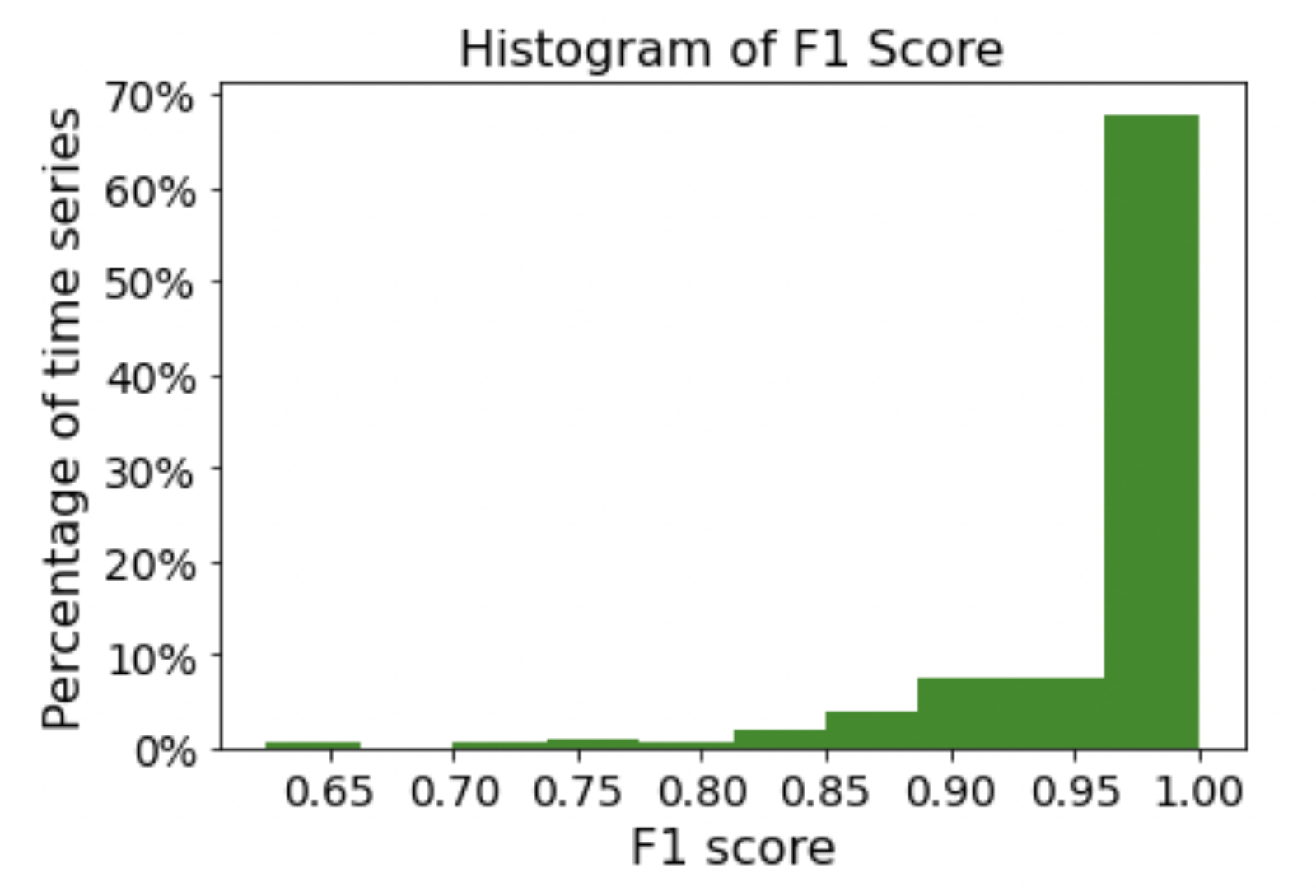} 
    \caption{Distribution of mSelect F1 scores}
    \label{fig:mselect_f1score}
\end{figure}

\begin{figure}[htbp]
    \centering
    \includegraphics[width=0.9\linewidth]{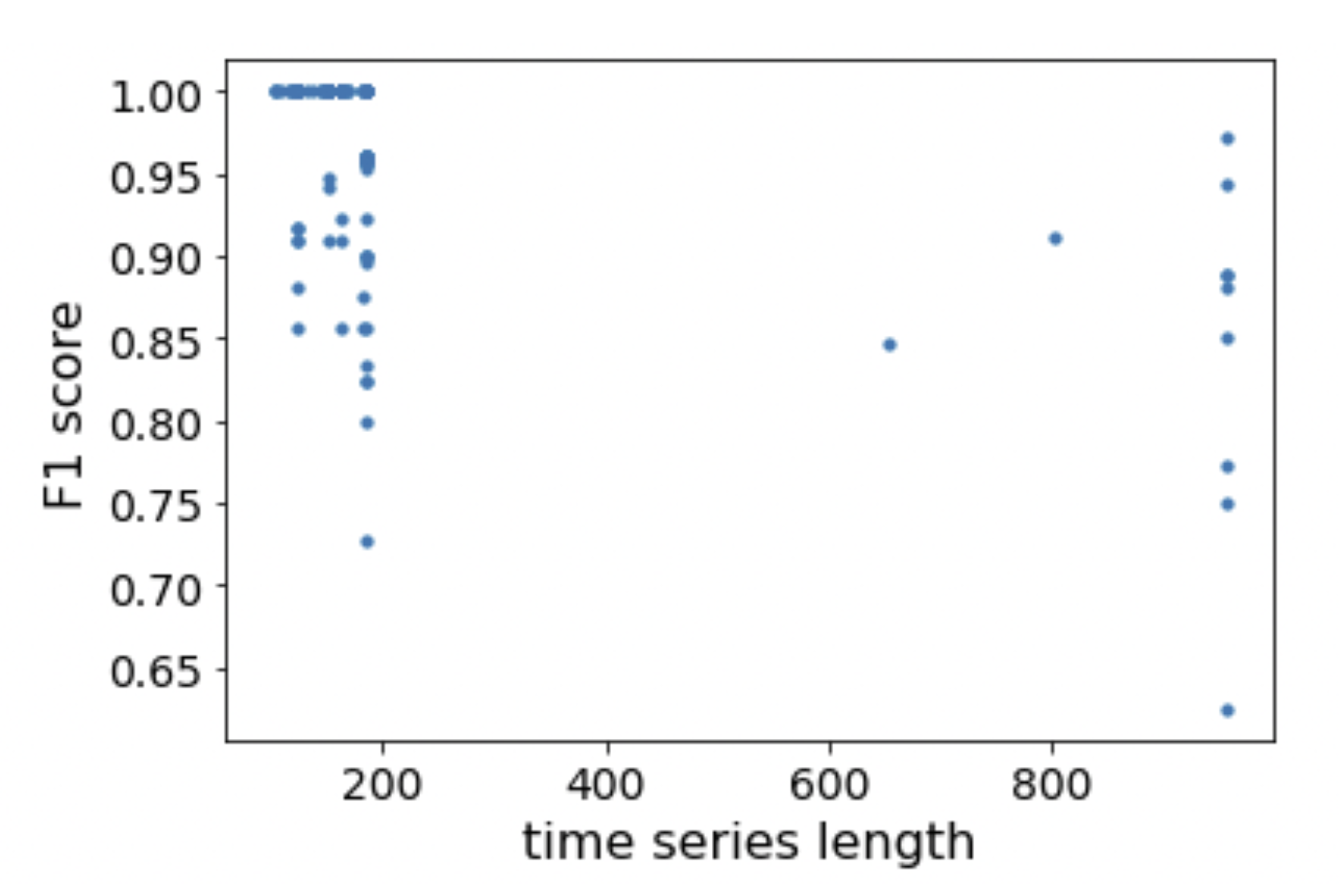} 
    \caption{mSelect F1 score distribution across time series of varying lengths}
    \label{fig:mselect_length}
\end{figure}

We evaluated the F1 score results of mSelect on 205 time series. Approximately 67\% of the time series achieved an F1 score between 0.95 and 1.00, as shown in Figure~\ref{fig:mselect_f1score}. mSelect showed superior performance on shorter time series, while for time series longer than 800, the F1 score ranged from 0.75 to 0.97, as illustrated in Figure~\ref{fig:mselect_length}.

\clearpage
\let\cleardoublepage\relax
\end{document}